\documentclass[10pt,twocolumn,letterpaper]{article}

\usepackage[pagenumbers]{cvpr}              

\usepackage[linesnumbered,ruled,vlined]{algorithm2e}
\usepackage[accsupp]{axessibility}

\usepackage{multirow}
\usepackage{multicol}
\usepackage{graphicx}
\usepackage{booktabs}
\usepackage{array}
\usepackage{makecell}

\definecolor{cvprblue}{rgb}{0.21,0.49,0.74}
\usepackage[pagebackref,breaklinks,colorlinks,allcolors=cvprblue]{hyperref}

\def\paperID{11336} 
\def\confName{CVPR}
\def\confYear{2025}

\title{CoMatcher: Multi-View Collaborative Feature Matching}

\author{
Jintao Zhang$^{1}$, Zimin Xia$^{2}$\thanks{Corresponding author.}, Mingyue Dong$^{1}$, Shuhan Shen$^{3}$, Linwei Yue$^{4}$, Xianwei Zheng$^{1\ast}$ \\
$^{1}$The State Key Lab. LIESMARS, Wuhan University \\
$^{2}$École Polytechnique Fédérale de Lausanne (EPFL) \\
$^{3}$Chinese Academy of Sciences \quad $^{4}$China University of Geosciences
}

\begin{document}
\maketitle
\begin{abstract}
This paper proposes a multi-view collaborative matching strategy for reliable track construction in complex scenarios.
We observe that the pairwise matching paradigms applied to image set matching often result in ambiguous estimation when the selected independent pairs exhibit significant occlusions or extreme viewpoint changes.
This challenge primarily stems from the inherent uncertainty in interpreting intricate 3D structures based on limited two-view observations, as the 3D-to-2D projection leads to significant information loss.
To address this, we introduce CoMatcher, a deep multi-view matcher to (i) leverage complementary context cues from different views to form a holistic 3D scene understanding and (ii) utilize cross-view projection consistency to infer a reliable global solution.
Building on CoMatcher, we develop a groupwise framework that fully exploits cross-view relationships for large-scale matching tasks.
Extensive experiments on various complex scenarios demonstrate the superiority of our method over the mainstream two-view matching paradigm.
Code is available at \url{https://github.com/EATMustard/CoMatcher}.
\end{abstract}
\section{Introduction}
\label{sec:intro}

\begin{figure}
    \centering\includegraphics[width=1\linewidth]{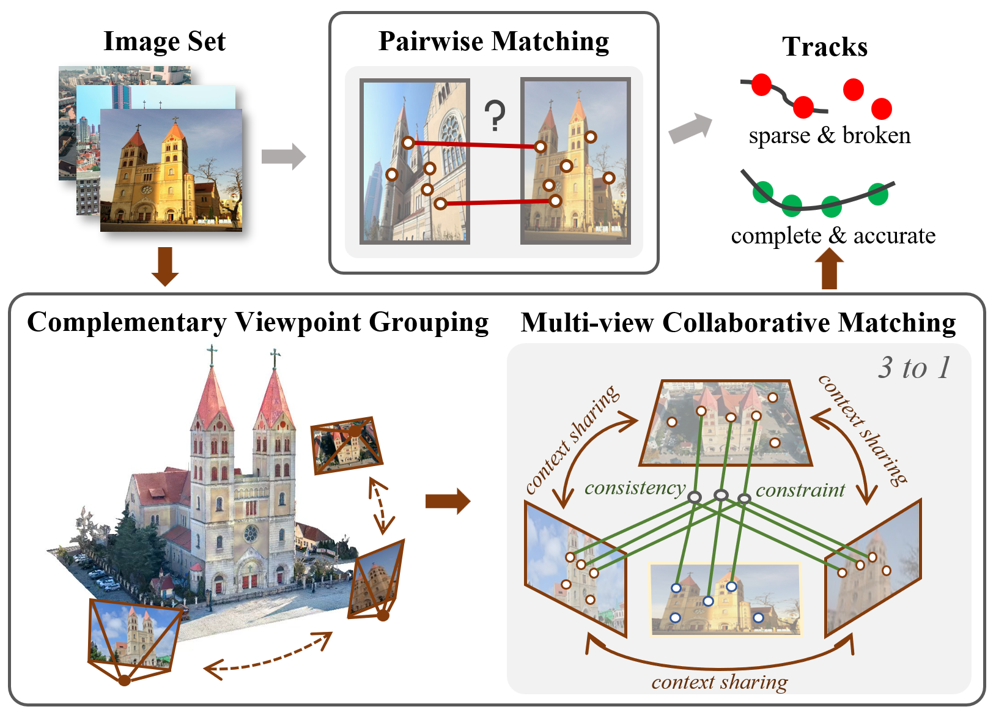}
    \caption{
    \textbf{From two view to multi-view.} Unlike pairwise schemes prone to uncertainty, we first partition the image set into co-visible groups and collaboratively estimate group-wide correspondences using a reliable deep multi-view matcher. Benefiting from holistic scene understanding and consistency constraint, our method generates high-quality tracks in complex scenarios.
}
    \label{fig1}
\end{figure}

Research on recovering 3D structures and camera poses from multiple views storing scene information has a long history~\cite{hartley2003multiple, jared2015reconstructing, mur2015orb}. 
A core component of this process is feature matching, which aims to estimate point correspondences across the image set~\cite{ma2021image, xu2024local, jin2021image}.
For computational flexibility, existing methods often decompose the set into pairs of co-visible images~\cite{agarwal2009building, frahm2010building} and apply a two-view matching approach to each pair independently~\cite{lowe2004distinctive, sarlin2020superglue}.
The pairwise matches are then merged into comprehensive multi-view tracks~\cite{agarwal2009building, schonberger2016structure}. 
Following this pairwise framework to build systems for localization~\cite{sarlin2019coarse, humenberger2020robust} and mapping~\cite{schonberger2016structure, lindenberger2021pixel} has become a entrenched research paradigm.

Recent advances in learning-based two-view matchers have greatly improved performance in most cases~\cite{sarlin2020superglue, lindenberger2023lightglue, sun2021loftr, edstedt2024roma}. 
Inspired by Transformer~\cite{vaswani2017attention}, these methods employ deep networks to jointly analyze global context from both views. 
This enables models to learn to infer underlying 3D scenes by integrating spatial and visual cues, which has been proven crucial for accurate estimations~\cite{sarlin2020superglue, leroy2024grounding}.

Despite great progress, current two-view matchers still struggle in challenging wide-baseline scenarios, particularly with severe occlusions and repetitive textures~\cite{jin2021image}.
We argue that one potential cause for this difficulty lies in the inherent uncertainty of interpreting complex 3D structures from limited two-view observations— a task that even humans find prone to perceptual ambiguity.
Projecting complex 3D geometries onto 2D planes inevitably results in significant loss of scene information.
For example, spatially distant regions may appear close in 2D.
This makes it highly unreliable to infer the original scene solely from ambiguous two-view context~\cite{cai2023doppelgangers, xiao2024spatialtracker}.
Moreover, while matchers can learn two-view geometric priors, this alone is also inadequate—capturing abrupt depth discontinuities requires stronger constraints~\cite{leroy2024grounding, barroso2024matching}.
Errors in pairwise correspondences compound during merging, significantly impacting downstream tasks, especially in large-scale datasets.

To address these challenges, we propose a shift in focus: instead of optimizing two-view matchers, a more effective strategy is to directly leverage the rich relationships inherent in raw multi-view observations.
Building on this, we introduce a unified formulation that collaboratively establishes correspondences between a complementary view group and a target view.
This approach fosters holistic scene understanding by integrating contextual information from multiple views, while multi-view geometric priors provide stronger constraints for reliable reasoning.
Moreover, matches across views naturally adhere to cross-view projection consistency.
This facilitates a mutual verification process, ensuring confident inference of a global solution.

Inspired by existing two-view methods~\cite{sarlin2020superglue, sun2021loftr}, we develop CoMatcher, a deep-learning architecture for multi-view feature matching.
Context sharing is achieved through a GNN with multi-view receptive fields.
While a larger feature space enriches context, it also introduces noise, complicating learning.
To address this, we propose to constrain the search space for each point by leveraging cross-view projection geometry.
Furthermore, we redesign the feature correlation layers in cross-attention to enable consistent multi-view reasoning.
This design exploits the property that correlation scores encode both reasoning outcomes and confidence, enabling progressive layer-by-layer integration of multi-view estimates into a globally consistent solution.

To scale CoMatcher for large-scale tasks like SfM~\cite{schonberger2016structure, pan2024global}, we develop a groupwise matching pipeline (see Fig.~\ref{fig1}).
It introduces a novel set partitioning algorithm that uses co-visibility information to group images, followed by group-level matching using CoMatcher.
By preserving cross-view relationships, our framework achieves comprehensive improvements over pairwise methods in extensive experiments on wide-baseline scenarios.
The main contributions are summarized as follows:
(i) A novel multi-view matching paradigm, comprising a deep collaborative matcher and a scalable groupwise pipeline, designed to overcome the inherent uncertainty of pairwise approaches. 
(ii) A multi-view representation learning network architecture that enables holistic scene understanding.
(iii) A multi-view feature correlation strategy that integrates uncertain individual estimates into a globally consistent solution.
\section{Related work}
\label{sec:related work}

\textbf{Two-view matching} is a fundamental problem in computer vision. 
The most prevalent paradigm involves sparsifying images into keypoints, each represented by a high-dimensional vector encoding its visual context~\cite{ma2021image, jin2021image}.
This transforms matching into a search problem, with match likelihood expressed by the inner product of corresponding vectors~\cite{lowe2004distinctive}.
Modern methods leverage deep networks for representation learning to extract feature points~\cite{yi2016lift, detone2018superpoint, revaud2019r2d2, dusmanu2019d2, tyszkiewicz2020disk, edstedt2024dedode}, outperforming traditional hand-crafted encodings~\cite{rublee2011orb, lowe2004distinctive, bay2006surf} and achieving significant advancements on mainstream benchmarks~\cite{balntas2017hpatches, jin2021image, schonberger2017comparative}. 
In parallel, other works aim to address the limitations of heuristic matching strategies by leveraging networks with global modeling capabilities to directly learn the matching process~\cite{sarlin2020superglue, lindenberger2023lightglue, jiang2024omniglue, chen2021learning}.
Such data-driven methods can effectively capture priors of the underlying 3D scene, which are essential for accurate estimation.
Unlike keypoint-based approaches, another typical method retains all pixels for dense inference~\cite{sun2021loftr, edstedt2024roma, chen2022aspanformer, truong2021learning, leroy2024grounding}.
While this has led to notable progress in weakly textured regions, such methods require greater computational resources~\cite{wang2024efficient, lindenberger2023lightglue} and struggle with multi-view inconsistency~\cite{he2024detector}.
In summary, while these methods perform well in most scenarios, they remain constrained in complex wide-baseline settings due to the limited receptive field inherent to two-view systems.

\textbf{Multi-view matching} focuses on establishing correspondences across a larger-scale image collection.
A typical pairwise matching approach decomposes the problem into two-view matching instances and subsequently merges the pairwise results using multi-view projection consistency~\cite{agarwal2009building, schonberger2016structure, sarlin2019coarse, sattler2017large}.
During merging, some works explore how consistency assumptions can filter and refine coarse matches~\cite{maset2017practical, sun2023unified}, while others focus on efficiency, developing better strategies to reduce redundant matching computations~\cite{olsson2011stable, zhang2010efficient}.
However, as a post-processing step, merging cannot address catastrophic failures that occur during the matching stage.
Another line of work treats matching as a point tracking problem~\cite{doersch2022tap, harley2022particle,doersch2023tapir, karaev2024cotracker}, using optical flow-like frameworks to reason about the positions of query pixels across multiple images.
While this approach is well-suited for video inputs, its effectiveness on unstructured, wide-baseline image collections remains largely unverified.
The most relevant method to ours is~\cite{roessle2023end2end}, which performs end-to-end multi-view matching and pose estimation.
However, it is limited in scalability, as it can only handle a small image group.
In contrast, our framework is highly flexible, with the potential to scale to an arbitrary number of images.

\section{Methodology}
\label{sec:method}

\textbf{Formulation.}
Given a set of $N_{I}$ images, \(\mathcal{I} = \{ I_{i} \mid i = 1, \ldots, N_{I} \}\), of a scene, the objective of feature matching is to identify the 2D position coordinates $\mathcal{M}_{k} = \{ \mathbf{p}^{I_{i}} \}$ in $\mathcal{I}$ that correspond to identical 3D scene points \(\mathcal{X} = \{ \mathbf{x}_{k} \mid k = 1, \ldots, N_{X} \}\), where \(\mathbf{p}^{I_{i}} \in \mathbb{R}^{2}\) are the pixel coordinates in $I_{i}$, and each $\mathcal{M}_{k}$ is called a \textit{track}.

\noindent\textbf{Motivation.}
\begin{figure*}[h]
    \centering
    \includegraphics[width=0.95\linewidth]{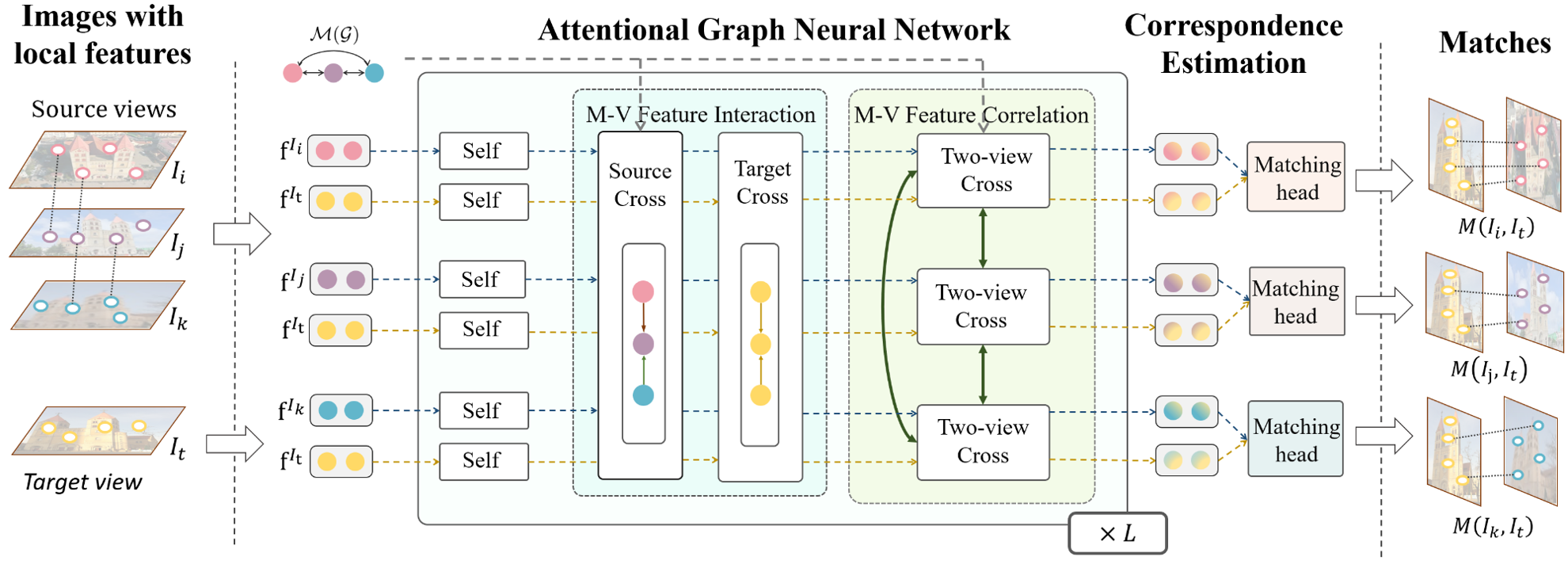}
    \caption{\textbf{The CoMatcher architecture} can be viewed as an extension of parallel two-view matchers, enhanced with two core components: the multi-view (M-V) feature interaction module (Sec.~\ref{sec:multi-view feature interaction}) and the multi-view (M-V) feature correlation strategy (Sec.~\ref{sec:multi-view feature correlation}).The first component employs multi-view cross-attention to enhance the representation of all source and target view features, while leveraging the cross-view projection geometry from $\mathcal{M}(\mathcal{G})$ to constrain the point search space. The second component aggregates correlation scores from multiple two-view cross-attention modules based on estimated confidence, guiding the network to achieve multi-view consistent reasoning.}
    \label{fig2}
\end{figure*}
Constructing complete tracks requires estimating correspondences between each image \( I_{t}\subset \mathcal{I} \) and its co-visible image set \( \mathcal{C}_{t} \).
While the pairwise approach offers implementation flexibility for this problem, its reliance on two-view inputs in a single estimation often leads to uncertainty in complex scenes.
In practice, images within \( \mathcal{C}_{t} \) are often not independent—most are captured from adjacent positions but with differing viewing angles.
This inspires us to group \( \mathcal{C}_{t} \) based on image correlations, where each complementary group offers a mapping from multiple planar views to a 3D sub-scene.
In the image matching problem, the correspondence between \( I_{t} \) and a group must adhere to certain 2D-to-3D physical constraints, such as cross-view projection consistency.
Rather than independently estimating and then integrating them with constraints, it is more efficient to unify this process and learn underlying priors directly from the data.
Thus, we formulate CoMatcher to address the $1$-to-$N$ matching problem (see Fig.~\ref{fig2}).

\noindent\textbf{Groupwise matching pipeline.}
Based on the above, we propose a three-step pipeline for constructing tracks: grouping, connecting, and matching.
First, we treat the original set of views $\mathcal{I}$ as a union of smaller image groups $\{\mathcal{G}_{s} \mid s=1,  \ldots,N_{G}\}$, each representing a localized scene.
This involves a pre-processing step to partition the set. 
The most critical aspect of this is determining the correlation between two images, i.e., whether they describe the same structure.
Building on existing research~\cite{arandjelovic2016netvlad, radenovic2018fine, sivic2003video}, we employ heuristic techniques to design a grouping algorithm.
Next, although CoMatcher can directly learn multi-view geometry, certain fine-grained physical rules are challenging to model.
Therefore, we explicitly establish relationships within each group to guide subsequent matching.
For each group, we construct tracks using existing frameworks, which provide a set of 3D points along with their multi-view projections and implicitly encode camera poses.
These cues are incorporated as inputs to the network to steer the inference process.
Finally, each group of images is treated as a whole to collaboratively match with other images in $\mathcal{I}$ using CoMatcher.
Additional details on grouping and the overall framework are provided in the supplementary materials.

\subsection{CoMatcher overview}
As shown in Fig.~\ref{fig2}, CoMatcher, as a deep sparse matcher, estimates point correspondences between each \textit{source} view in a group $I_{i} \in \mathcal{G}$ and a \textit{target} view $I_{t}$.
The network takes as input the extracted local features $\{(\mathbf{p}_{k}^{I_{i}}, \mathbf{d}_{k}^{I_{i}}) \mid k=1, \ldots, N_{F}\}$ from the $M$ source views and the target view, along with the precomputed tracks $\mathcal{M}(\mathcal{G})$ of $\mathcal{G}$.
Each local feature consists of keypoint locations $\mathbf{p}_{k}$ and their corresponding visual descriptors $\mathbf{d}_{k}$.
Initially, the target view features are broadcast and paired with those of each source view to establish the initial states.
These $M$ feature pairs are then refined using a Graph Neural Network (GNN) with alternating self-attention and cross-attention modules, repeated $L$ times.
Finally, the enhanced features are fed into matching heads to infer the final correspondences.

The design of CoMatcher revolves around two core questions: (i) how can multi-view context be leveraged to learn better point representations (Sec.~\ref{sec:multi-view feature interaction}), and (ii) how can the estimations across multiple views be constrained to satisfy cross-view consistency (Sec.~\ref{sec:multi-view feature correlation})?
After thoroughly discussing these questions, we introduce the prediction heads (Sec.~\ref{sec:correspondence prediction}) and the loss function (Sec.~\ref{sec:loss}).

\subsection{Multi-view feature interaction}
\label{sec:multi-view feature interaction}
Consider an occluded scene. 
When observing such structures from a frontal view, points near occlusion boundaries are often contaminated by irrelevant context. 
This noise significantly compromises the quality of point features, creating substantial challenges for matching.
Multi-view learning provides a promising solution to this issue.
For these ambiguous point features, it can integrate observations of the same area from multiple other viewpoints to refine their representations. 
This involves a querying process, which establishes a multi-view receptive field for each point.

Given a set of co-visible source views, CoMatcher leverages a multi-view cross-attention mechanism to aggregate these context (Source Cross).
For a point $u$ in source view $I_{i}$, we attend to all points in the other source views:
{\small
\begin{align}
    \mathbf{m}_{u}^{I_{i} \leftarrow \mathcal{W}_j} = \frac{1}{M-1} \sum_{I_{j} \in \mathcal{G} \setminus I_{i}} \sum_{v \in \mathcal{W}_{j}} \operatorname{Softmax} \left(a_{u v}^{I_{i} I_{j}}\right) \mathbf{v}_{v}^{I_{j}}.
\end{align}
}
Here, $\mathcal{W}_{j}$ denotes the point set in the source view $I_{j}$, while $a_{u v}^{I_{i} I_{j}}$ represents the similarity score computed by treating point $u$ in $I_{i}$ as the query and point $v$ in $I_{j}$ as the key. 
$\mathbf{v}_{v}^{I_{j}}$ is the value of point $v$.
$\mathbf{m}_{u}^{I_{i}}$ represents the message vector from $\mathcal{W}_{j}$ to $u$ of the GNN~\cite{sarlin2020superglue}.
We uniformly aggregate features from each source view, encouraging the model to integrate multi-view information comprehensively rather than overemphasizing images from similar perspectives.

However, for a point near an occlusion boundary in one source view, how does it identify the corresponding points observing the same location in other source views? 
This implies a matching process. 
While the network could directly learn this, it would introduce additional complexity, as establishing correspondences between source views is not our primary objective.
Moreover, while multi-view setups provide abundant contextual information, much of it is irrelevant, such as regions lacking co-visibility. 
This redundancy also imposes an additional burden on learning.

To address this, we propose a geometric constraint mechanism to explicitly guide the attention range of each point.
By incorporating relative positional encoding into the attention score computation, we enable the scores to depend on two critical factors: feature correlation and the geometric distance $\Delta\mathbf{p}$ between two points:
\begin{align}
a_{u v}^{I_{i} I_{j}} = \left(\mathbf{q}_{u}^{I_{i}}\right)^{\top} \mathbf{R}\left[\Delta\mathbf{p}_{u v}^{I_{i} I_{j}}\right] \mathbf{k}_{v}^{I_{j}}.
\end{align}
Here, $\mathbf{R}\left[\cdot\right]$ is the rotation encoding matrix~\cite{su2024roformer}.

While the 2D relative distance between two points within a single image is well-defined, across different views, it depends on the projection transformation.
We use the precomputed tracks $\mathcal{M}(\mathcal{G})$ to find the projection position of   $\mathbf{p}_u^{I_{i}}$ on $I_j$, which can be denoted as  $\mathbf{p}_{w}^{I_{j}}$. When no corresponding point in $I_j$ can be found for $\mathbf{p}_u^{I_{i}}$, we simply assign the relative position to $\boldsymbol{0}$ so that the attention score depends solely on the feature similarity. The calculation of the relative position can be formulated as follows:
\begin{align}
\Delta\mathbf{p}_{u v}^{I_{i} I_{j}}=
    \begin{cases}
        \mathbf{p}_{w}^{I_{j}} - \mathbf{p}_{v}^{I_{j}} & {(u,w) \in M(I_{i},I_{j})} \\
       \boldsymbol{0} & {(u,\emptyset) \in M(I_{i},I_{j})}.
    \end{cases}
\end{align}

Additionally, we also interact with all target view features after source cross-attention (Target Cross).
For a point $t$ in view $I_{t}$ paired with source view $I_{i}$, we attend to the same point $t$ across the remaining $M-1$ target views.
This is analogous to operations in some works that aggregate features along the temporal dimension~\cite{doersch2023tapir,karaev2024cotracker}.
Unlike the approach for source views, we compute the attention scores solely based on feature similarity.
Despite its low computational complexity, this module complements the Source Cross, mutually enhancing each other to help the network implicitly infer globally optimal correspondences.

\subsection{Multi-view feature correlation}
\label{sec:multi-view feature correlation}
For a track in a source view set originating from a 3D point, its corresponding 2D keypoints $\left\{ \mathbf{p}_{u}^{I_{i}},\mathbf{p}_{v}^{I_{j}}, \dots \right\}$ should exhibit consistent matches in the target view—either corresponding to the same single point or having no corresponding point at all.
This physical constraint is often utilized by existing methods as a post-processing step to filter out erroneous results when merging multiple two-view correspondences.
However, as a multi-view matcher, we aim for CoMatcher to inherently satisfy this constraint in its reasoning, thereby enhancing its reliability and confidence.

\begin{figure}
    \centering
\includegraphics[width=0.95\linewidth]{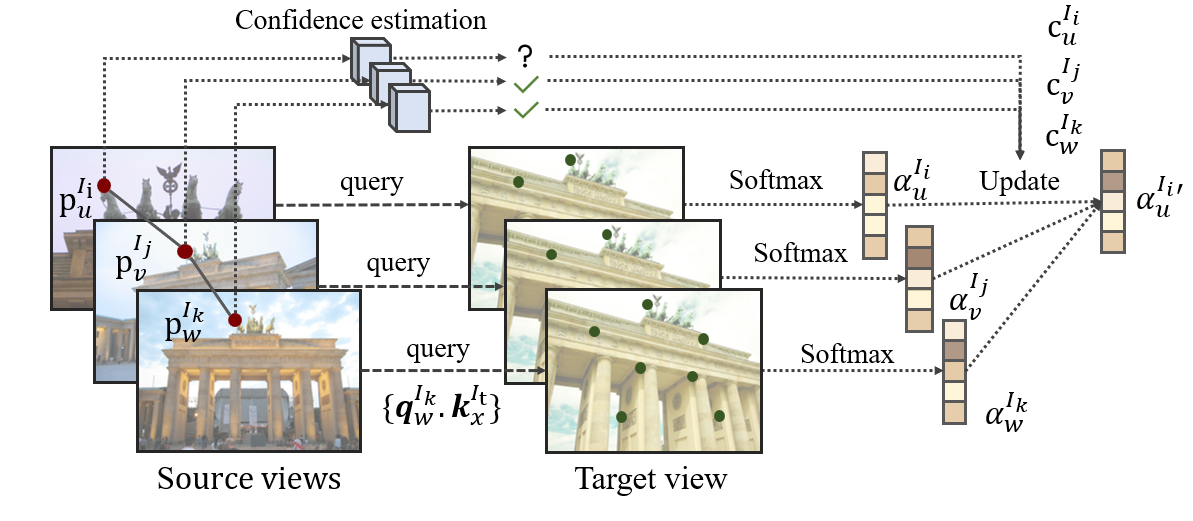}
    \caption{\textbf{The multi-view feature correlation strategy} consists of two steps: we first identify potentially erroneous points using a confidence estimator, and then correct their estimation by aggregating attention correlation vectors from other views.}
    \label{fig4}
\end{figure}

Achieving this involves two key processes:
(i) identifying points that may lead to errors in the \textit{early} layers of the network, and
(ii) promptly guiding these points toward correct estimation using inference information from other views.
To accomplish this, we propose a two-step multi-view feature correlation strategy (see Fig.~\ref{fig4}).

First, to identify ambiguous points in the source views, we utilize a lightweight head at the end of each layer to predict the confidence of each point:
\begin{align}
c_{u}^{I_{i}}=\operatorname{Sigmoid}\left(\operatorname{MLP}\left(\mathbf{f}_{u}^{I_{i}}\right)\right) \in[0,1].
\end{align}
Here, $\operatorname{MLP}$ represents a multi-layer perceptron and $\mathbf{f}_{u}^{I_{i}}$ is the feature of $u$.
This confidence score reflects the assessment of each point in its current state: higher values indicate that, at this layer, the point is either converging toward a reliable match or confidently identified as having no match.

In each layer, we define a distinct hyperparameter threshold $\theta$. Points with confidence scores below this threshold are identified as potentially ambiguous and likely to lead to incorrect inference.
We progressively increase the value of $\theta$ in later layers, as the confidence of overall inference generally improves with each successive layer.

Next, we redesign the parallel two-view cross-attention modules to correct the erroneous estimations of ambiguous points.
In these modules, each point in the source view computes an attention distribution vector to query~\cite{vaswani2017attention}:
\begin{equation}
\boldsymbol{\alpha}_{u}^{I_{i}} = \underset{x \in \mathcal{W}_{t}}{\operatorname{Softmax}}(a_{u x}^{I_{i} I_{t}}) .
\end{equation}
This vector represents the degree to which the point attends to in view $I_{t}$, with a peak indicating potential matching regions of interest.
For ambiguous points, this similarity is not reliable, often leading to incorrect feature aggregation.
Assuming that point $u$ in the source view $I_{i}$ is an ambiguous point, we update its original attention distribution by applying a weighted average to the distributions of its corresponding points across multi-view:
{\small
\begin{align}
\boldsymbol{\alpha}_{u}^{I_{i} \prime} = c_{u}^{I_{i}} \boldsymbol{\alpha}_{u}^{I_{i}} + (1 - c_{u}^{I_{i}}) \frac{\sum_{v \in \mathcal{D}_{m}^{u}}  c_{v}^{I_{j}} \boldsymbol{\alpha}_{v}^{I_{j}}}{\sum_{v \in \mathcal{D}_{m}^{u}}  c_{v}^{I_{j}}} \quad \text{s.t. }c_{u}^{I_{i}} < \theta .
\end{align}
}

Here, $\mathcal{D}_{m}^{u}$ represents other points in the track.
We embed this strategy into the model inference process, which significantly enhances the reliability of estimation under challenging conditions despite its simple implementation.

\subsection{Correspondence prediction}
\label{sec:correspondence prediction}
As shown in Fig.~\ref{fig2}, a lightweight matching head operates in parallel for each pair of network output features to predict correspondences between two views~\cite{sarlin2020superglue, lindenberger2023lightglue}.
In each head, we begin by calculating a score matrix from feature correlations between two views: $\mathbf{S}(u, x)=\operatorname{L}(\mathbf{f}_{u}^{I_{i}}) \cdot \operatorname{L}(\mathbf{f}_{x}^{I_{t}})$, where $\operatorname{L}$ is the linear transform. Then we apply dual-softmax~\cite{tyszkiewicz2020disk, sun2021loftr, lindenberger2023lightglue} to derive the matching probabilities of two points: $
    \mathbf{S}'(u, x)= \operatorname{Softmax}(\mathbf{S}(u, \cdot))_{x} \cdot \operatorname{Softmax}(\mathbf{S}(\cdot, x))_{u}$.
Additionally, we employ the method proposed in~\cite{lindenberger2023lightglue} to calculate the matching probability for each point: $\sigma_{i}^{I_{i}}=\operatorname{Sigmoid}(\operatorname{L}(\mathbf{f}_{i}^{I_{i}}))$.
The final assignment matrix is a combination of the two probabilities:
\begin{equation}
    \mathbf{P}(u,x)=\mathbf{S}'(u, x)\sigma_{u}^{I_{i}}\sigma_{x}^{I_{t}}.
\end{equation}
Using mutual nearest neighbors and a threshold, we filter the matching results from the assignment matrix~\cite{sarlin2020superglue, sun2021loftr}.

\subsection{Loss}
\label{sec:loss}
CoMatcher supervises each source-target image pair individually, with the loss for each pairs composed of two components: correspondence and confidence:
\begin{equation}
\mathcal{L}_{\text{total}} = \frac{1}{M}\sum_{I_{i}\in \mathcal{G}}(\mathcal{L}_{\text{corr}}(I_{i},I_{t})+\alpha\mathcal{L_{\text{conf}}}(I_{i})).
\end{equation}

\noindent\textbf{Correspondence supervision.}
For each pair $I_{i}$ and $I_{t}$, we perform two-view transformations using relative poses or homography to compute ground truth labels~\cite{sarlin2020superglue, sun2021loftr, lindenberger2023lightglue}.
We minimize the negative log-likelihood of the assignment matrix, following~\cite{lindenberger2023lightglue}.
More details are provided in the supplementary materials.

\noindent\textbf{Confidence supervision.}
To train the confidence estimators at each layer, we define the point confidence as the consistency probability between its correspondence estimated at the current layer and the final estimation~\cite{elbayad2020depth, schuster2022confident, lindenberger2023lightglue}. At each layer, predictions are obtained via dual-softmax on current features, without introducing additional matching probabilities or parameters. The ground truth label $y$ indicates whether the two estimations are consistent. This is supervised using a binary cross-entropy loss:
\begin{equation}
\mathcal{L_{\text{conf}}}(I_{i})=\frac{1}{L-1}\sum_{\ell}\sum_{u \in \mathcal{W}_{i}}\operatorname{CE}({ }^{\ell} c_{u}^{I_{i}}, { }^{\ell} y_{u}^{I_{i}}),
\end{equation}
where $\ell \in\{1, \ldots, L-1\}$.

\section{Experiments}
\label{sec:experiments}
In this section, we first introduce the datasets used, followed by our implementation details.
Then, our CoMatcher network is compared to previous state-of-the-art baselines for homography and camera pose estimation.
Next, we integrate CoMatcher into groupwise matching framework and evaluate it against pairwise method on a large-scale benchmark.
Finally, an extensive ablation study is provided.

\subsection{Datasets}
\begin{figure*}
  \centering
   \includegraphics[width=0.953\linewidth]{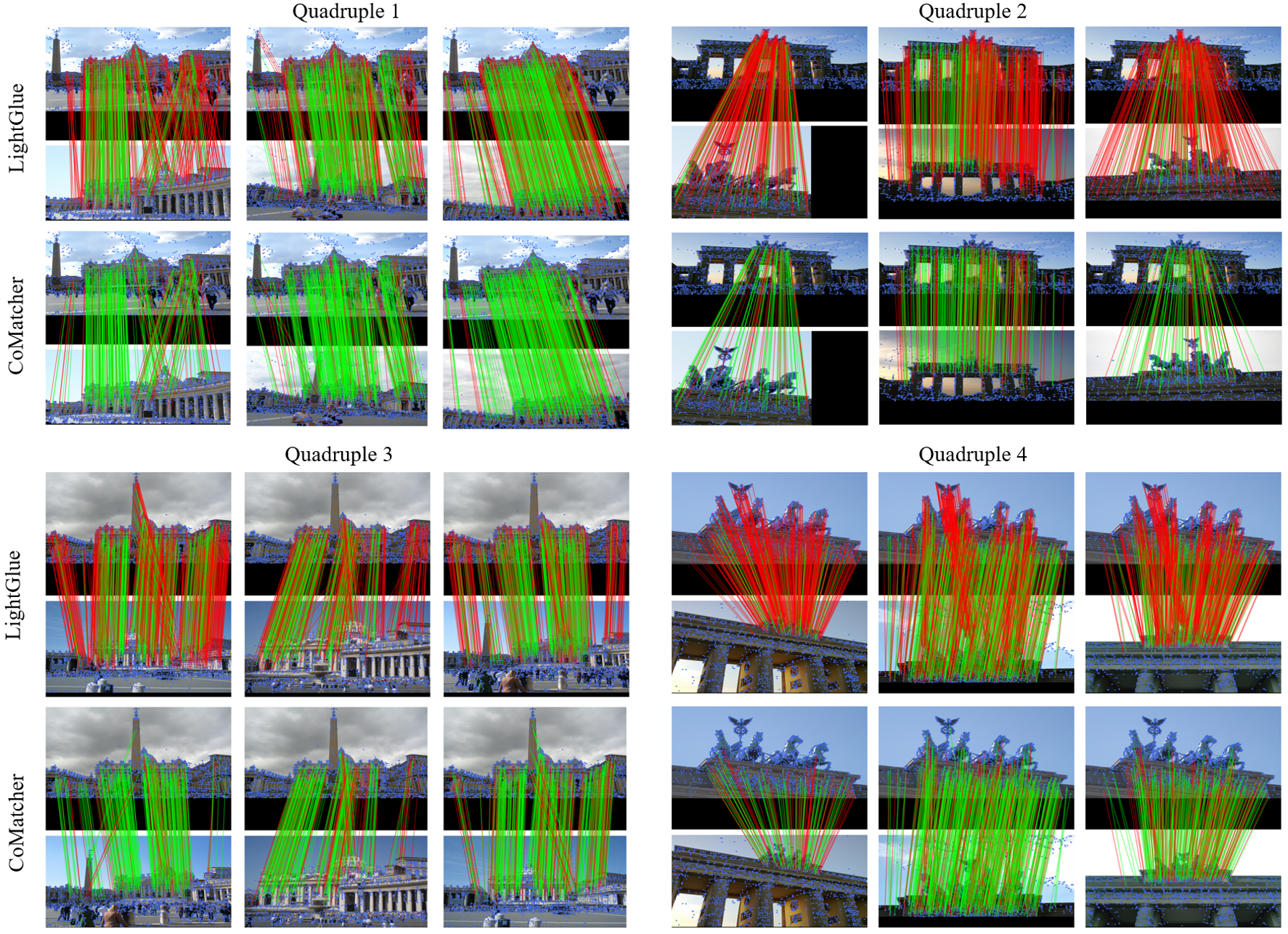}
   \caption{\textbf{Qualitative comparison on MegaDepth.} For each quadruple with a target view (top), correspondences (green for correct, red for incorrect) predicted by CoMatcher and LightGlue are shown.  
Using identical local features as input, CoMatcher achieves significantly more \textit{reliable} estimations, even at challenging semantic edges with depth discontinuities. 
This stems from its holistic scene understanding through multi-view cues to effectively address occlusions (Q3), large-scale variations (Q2, 4), and repetitive textures (Q1, 2, 3, 4).}
   \label{fig5}
\end{figure*}
The \textbf{HPatches} dataset is used for homography estimation~\cite{balntas2017hpatches}. 
It consists of 116 strictly planar scenes, each containing 6 images with variations in viewpoint and illumination. 
For camera pose estimation tasks, we utilize the \textbf{MegaDepth} dataset, an outdoor dataset that exhibits strong occlusions and significant structural changes~\cite{li2018megadepth}. 
We selected two scenes, ``Sacre Coeur'' and ``St. Peter's Square'', from which 1500 co-visible quadruplets were sampled in a way that balances difficulty based on visual overlap~\cite{sun2021loftr, lindenberger2023lightglue, wang2024efficient}.
The \textbf{Image Matching Challenge 2020} benchmark provides a comprehensive evaluation protocol, including datasets that cover multiple challenging outdoor wide-baseline scenes. 
From the phototourism dataset, we selected 3 validated scenes, each consisting of about 100 images.
Additionally, we conducted an ablation study on a synthetic homography dataset~\cite{radenovic2018revisiting} used for training.

\subsection{Implementation details}
\label{sec:implemetation details}
CoMatcher is trained in two steps, following~\cite{sarlin2020superglue, lindenberger2023lightglue}. 
We first pre-train the model on a large-scale synthetic homography dataset~\cite{radenovic2018revisiting}, leveraging noise-free homography for ground truth.
Next, fine-tuning is performed on the MegaDepth dataset~\cite{li2018megadepth} using the camera poses recovered by SfM as ground truth.
We sample 200 co-visible multi-view groups per scene and randomly select one image as the target view.
The size of source views is set to $M=4$ during training. 
The training process is carried out on two NVIDIA GeForce 4090 GPUs, taking about 6 days in total.
More details can be found in the supplementary materials.

\subsection{Homography estimation}
\label{sec:homography estimation}
CoMatcher is compared with three types of baselines. 
First, we include state-of-the-art (SOTA) sparse two-view matchers: nearest-neighbor with mutual check~\cite{lowe2004distinctive}, SuperGlue~\cite{sarlin2020superglue}, SGMNet~\cite{chen2021learning}, and LightGlue~\cite{lindenberger2023lightglue}, each paired with different feature extractors~\cite{lowe2004distinctive, detone2018superpoint, tyszkiewicz2020disk}. 
Second, we evaluate the multi-view matcher End2End~\cite{roessle2023end2end}.
For reference, we also compare against typical dense matchers LoFTR~\cite{sun2021loftr} and PDC-Net~\cite{truong2021learning}.
For CoMatcher and End2End, we use a single forward pass to obtain matches across all five image pairs, while others follow a pairwise approach.
To ensure a fair comparison, we adopt the setups of~\cite{lindenberger2023lightglue, sun2021loftr} for the number of features and image resizing.
Homography accuracy is evaluated using both robust (RANSAC~\cite{fischler1981random}) and non-robust (weighted DLT~\cite{hartley2003multiple}) solvers.
For each pair, we compute the mean reprojection error of the four image corners and report the area under the cumulative error curve (AUC) up to 1px, 3px, and 5px.

Tab.~\ref{tab1} shows that CoMatcher yields more accurate estimates than all sparse two-view matchers, highlighting the advantage of inference in multi-view feature space.
However, despite End2End also jointly reasoning over multiple views like ours, its performance is notably inferior.
This underscores the superiority of a $N$-to-$1$ architecture over $N$-to-$N$ for multi-view representation learning.
When compared to LoFTR, we observe slightly lower performance under low thresholds.
We attribute this mainly to the limitations of the feature extractor in keypoint non-repeatability and localization errors~\cite{lindenberger2021pixel}.
Additionally, DLT achieves accuracy close to RANSAC on most metrics, reflecting the high-quality correspondences from CoMatcher.

\subsection{Relative pose estimation}
\label{sec:relative pose estimation}
\begin{table}
\renewcommand{\arraystretch}{1.24} 
\centering
\small 
\begin{tabular}{p{0.4cm} @{\hskip 5pt}p{2.4cm} r@{\hskip 10pt}r} 
\toprule
\multicolumn{2}{c}{\multirow{2}{*}{Method}}                                                            & AUC - DLT            & AUC - RANSAC         \\ 
\cline{3-4}
\multicolumn{2}{c}{}                                                                                   & \multicolumn{2}{c}{@1px / @3px / @5px}     \\ 
\hline
\multirow{2}{*}{\rotatebox{90}{dense}}       & LoFTR          & \textbf{38.5} / 66.0 / 71.4   & \textbf{40.7} / 68.3 / 78.5  \\ 
                                                                                      & PDC-Net        & 36.0 / 65.3 / 73.0  & 37.9 / 67.6 / 77.4  \\ 
\hline
\multirow{7}{*}{\rotatebox{90}{sparse 2-view}}     & SIFT+NN+mutual & 0.0 / 0.0 / 0.0      & 35.9 / 65.0 / 75.6  \\
                                                                                      & SP+NN+mutual   & 0.0 / 1.9 / 3.4      & 34.8 / 64.1 / 74.8  \\
                                                                                      & SP+SuperGlue   & 32.2 / 65.1 / 75.7   & 37.2 / 68.0 / 78.7  \\
                                                                                      & SP+SGMNet      & 31.7 / 64.9 / 76.0   & 37.7 / 66.4 / 77.5  \\
                                                                                      & SP+LightGlue   & 35.4 / 67.5 / 77.7  & 37.2 / 67.8 / 78.1  \\
                                                                                      & DISK+NN+mutual & 1.8 / 5.2 / 7.8      & 37.9 / 58.0 / 68.3  \\
                                                                                      & DISK+LightGlue & 34.4 / 64.5 / 74.4   & 38.1 / 65.2 / 77.2  \\ 
\hline
\multirow{3}{*}{\rotatebox{90}{multi-view}} & SP+End2End     & 34.3 / 66.9 / 75.5   & 37.0 / 67.2 / 77.5  \\ 
                                                                                      & SP+CoMatcher   & 37.1 / \textbf{69.0} / \textbf{78.8}   & 38.4 / \textbf{68.9} / \textbf{79.0}  \\
                                                                                      & DISK+CoMatcher & 36.3 / 66.2 / 75.9   & 38.7 / 68.2 / 78.4  \\ 
\bottomrule
\end{tabular}
\caption{\textbf{Homography estimation on HPatches.} We report the area under the cumulative error curve (AUC) up to values of 1px, 3px and 5px, using DLT and RANSAC~\cite{fischler1981random} as homography solver.}
\label{tab1}
\end{table}

Next, we evaluate CoMatcher on challenging wide-baseline scenes, using the same baselines as in Sec.~\ref{sec:homography estimation}. 
For each quadruplet, including a selected target view, we still use a single forward pass for multi-view methods. 
Additionally, we test two newer dense methods, DUSt3R~\cite{wang2024dust3r} and MASt3R~\cite{leroy2024grounding}, following their evaluation setups for image resizing. 
All other settings remain consistent with~\cite{lindenberger2023lightglue}.
The essential matrix is estimated from the correspondences using both vanilla RANSAC~\cite{fischler1981random} and LO-RANSAC~\cite{chum2003locally}, following~\cite{lindenberger2023lightglue}.
The pose error is then computed as the maximum angular error of rotation and translation, derived from the decomposition of the essential matrix.
We report the AUC at $5^{\circ}$, $10^{\circ}$, and $20^{\circ}$, and record the average runtime of matching each quadruplet on a single 4090 GPU.

Tab.~\ref{tab2} shows that CoMatcher, as a sparse method, comprehensively enhances the estimation performance on different local features~\cite{detone2018superpoint} and~\cite{tyszkiewicz2020disk}.
Importantly, compared to other sparse two-view matchers like LightGlue,  a key strength of CoMatcher is its reliability, as illustrated in Fig.~\ref{fig4}.
This highlights the advantage of multi-view collaborative reasoning in understanding complex 3D structures, such as occlusions, even with suboptimal input local features in these areas.
Additionally, compared to End2End, CoMatcher achieves more accurate relative poses without the need for cumbersome end-to-end training.
While some dense methods may achieve better accuracy on certain metrics, their efficiency is generally much lower.  
By employing a space-for-time strategy to minimize inference steps, CoMatcher runs significantly faster than most methods.

\subsection{Evaluation on the IMC 2020 benchmark}
\label{sec:imc}
\begin{table}
\renewcommand{\arraystretch}{1.24} 
\centering
\small
\begin{tabular}{p{0.3cm} @{\hskip 3pt} l@{\hskip 3pt} c@{\hskip 3pt} c@{\hskip 3pt}c} 
\toprule
\multicolumn{2}{c}{\multirow{2}{*}{Method}} & AUC RANSAC & AUC LO-RANSAC & \multirow{2}{*}{\begin{tabular}[c]{@{}c@{}}Time\\ (ms)\end{tabular}} \\ 
\cline{3-4}
\multicolumn{2}{c}{} & \multicolumn{2}{c}{5° / 10° / 20°} & \\ 
\hline
\multirow{4}{*}{\rotatebox{90}{dense}}         & LoFTR         & 58.0 / 73.1 / 84.4 & 67.4 / 81.7 / 89.3 & 182 \\ 
                                               & PDCNet        & 54.7 / 73.1 / 83.5   & 67.1 / 80.2 / 87.0          & 231 \\ 
                                               & DUSt3R        & 42.4 / 56.7 / 64.2   & 58.1 / 70.4 / 79.6          & 264 \\ 
                                               & MASt3R        & 51.5 / 65.3 / 75.6   & 63.5 / 76.3 / 85.2          & 317 \\ 
\hline
\multirow{5}{*}{\rotatebox{90}{SuperPoint}}    & NN+mutual     & 35.3 / 58.3 / 53.7           & 51.4 / 67.3 / 75.9          & 9 \\ 
                                               & SuperGlue     & 55.8 / 72.8 / 84.1           & 65.1 / 77.2 / \textbf{89.2} & 87 \\ 
                                               & LightGlue     & 56.2 / 72.7 / 83.5           & 67.2 / 80.1 / 88.0          & 51 \\ 
                                               & End2End       & 55.3 / 71.4 / 81.2           & 67.4 / 81.5 / 87.0          & 152 \\ 
                                               & \textbf{CoMatcher}     & \textbf{57.2 / 73.9 / 84.8}  & \textbf{68.3 / 82.2} / 89.1 & 69 \\ 
\hline
\multirow{3}{*}{\rotatebox{90}{DISK}}          & NN+mutual     & 50.9 / 66.7 / 77.7           & 64.0 / 79.5 / 87.6          & 9 \\ 
                                               & LightGlue     & 53.2 / 69.2 / 80.2           & \textbf{68.6} / 80.4 / 87.2 & 54 \\ 
                                               & \textbf{CoMatcher}     & \textbf{54.9 / 71.2 / 82.0}  & 68.5 / \textbf{82.1 / 88.4} & 73 \\ 
\bottomrule
\end{tabular}
\caption{\textbf{Relative pose estimation on MegaDepth.} We report the AUC at $5^{\circ}$, $10^{\circ}$, and $20^{\circ}$ using different robust estimator, and the average runtime of matching each quadruplet.}
\label{tab2}
\end{table}

Next, we evaluate our groupwise framework against pairwise baselines~\cite{lowe2004distinctive, sarlin2020superglue, lindenberger2023lightglue} on the Image Matching Challenge (IMC) 2020 benchmark~\cite{jin2021image}.  
Given a large-scale unordered image collection, the benchmark requires providing matching results for all image pairs, which are then assessed on two downstream tasks: stereo and multi-view reconstruction.
For the stereo task, the accuracy of relative camera poses for each pair is evaluated from correspondences using RANSAC, as in Sec.~\ref{sec:relative pose estimation}.  
For the multi-view task, all correspondences are fed into COLMAP~\cite{schonberger2016structure} for SfM, with the final accuracy evaluated based on the estimated multi-view camera poses.
We report the AUC at $5^{\circ}$ and $10^{\circ}$ across both tasks.
Additionally, the average runtime is reported, which is the total matching time for the entire image set divided by the number of pairs.

Tab.~\ref{tab3} shows that our framework significantly outperforms existing pairwise approaches in both tasks.
This advantage primarily stems from our framework's ability to better leverage the relationships within the original image collection, enabling the generation of higher-quality tracks in such large-scale tasks.
Additional comparisons on track quality for SfM tasks are provided in the supplementary material.
In terms of efficiency, our approach is faster than SuperGlue, with the viewpoint grouping process accounting for only about 4\% of the total matching time.

\begin{table}
\renewcommand{\arraystretch}{1.2} 
\setlength{\tabcolsep}{4pt} 
\centering
\small
\begin{tabular}{p{0.4cm} l c c c} 
\toprule
\multicolumn{2}{c}{\multirow{2}{*}{Method}} & Stereo & Multi-View & \multirow{2}{*}{\makecell{Time\\(ms)}} \\ 
\cline{3-4}
\multicolumn{2}{c}{} & AUC 5°/10° & AUC 5°/10° & \\ 
\hline
\multirow{6}{*}{\rotatebox{90}{two-view}}   & SIFT+NN+mutual & 31.5 / 44.2    & 57.2 / 68.5        & 3 \\ 
                                            & SP+NN+mutual   & 28.6 / 40.3    & 52.9 / 63.4        & 3 \\ 
                                            & SP+SuperGlue          & 36.5 / 50.1    & 62.3 / 74.8        & 32 \\ 
                                            & SP+LightGlue          & 36.8 / 49.4    & 64.6 / 75.4        & 17 \\ 
                                            & DISK+NN+mutual & 35.4 / 47.4    & 59.3 / 70.2        & 3 \\ 
                                            & DISK+LightGlue        & 42.1 / 55.6 & 65.2 / 76.2 & 19 \\ 
\hline
\multirow{2}{*}{\rotatebox{90}{m-v}} & SP+\textbf{CoMatcher}   & 39.1 / 52.4    & 66.2 / 77.5 & 25 \\ 
                                     & DISK+\textbf{CoMatcher} & \textbf{44.9 / 57.3} & \textbf{67.1} / \textbf{78.4} & 29 \\ 
\bottomrule
\end{tabular}
\caption{\textbf{Evaluation on the IMC 2020 benchmark.}
We report the pose AUC at $5^{\circ}$, $10^{\circ}$ for two subtasks, stereo and multi-view reconstruction, along with the average matching runtime.}
\label{tab3}
\end{table}

\subsection{Ablation study}
\label{sec:ablation study}
\textbf{Understanding CoMatcher.}
The CoMatcher network introduces three key components: multi-view cross-attention modules, a geometric constraint mechanism, and a multi-view feature interaction strategy. We validate these design choices by evaluating on two benchmarks: the challenging synthetic homography dataset~\cite{radenovic2018revisiting}, where precision and recall metrics are employed, and HPatches~\cite{balntas2017hpatches}, where the AUC of homography estimation via weighted DLT is measured. For each image, 512 keypoints are extracted.

Tab.~\ref{tab4} shows the ablation results.
Excluding the two cross-attention modules for multi-view interaction reduces the model to a series of parallel two-view matchers~\cite{lindenberger2023lightglue, sarlin2020superglue}.
This simplification limits the model's ability to integrate multi-view context, leading to a significant performance decline.
Without the geometric constraint mechanism, the model performs even worse than in the previous case.
This suggests that direct feature aggregation from multi-view is challenging, as the network requires additional learning to distinguish meaningful context effectively.
Without multi-view feature correlation, the model loses its ability to guide inference for ambiguous points by leveraging consistency constraints.
This highlights the importance of guiding two-view feature correlation in the attention search process, a factor overlooked in previous work. 

\begin{figure}
  \centering
   \includegraphics[width=1\linewidth]{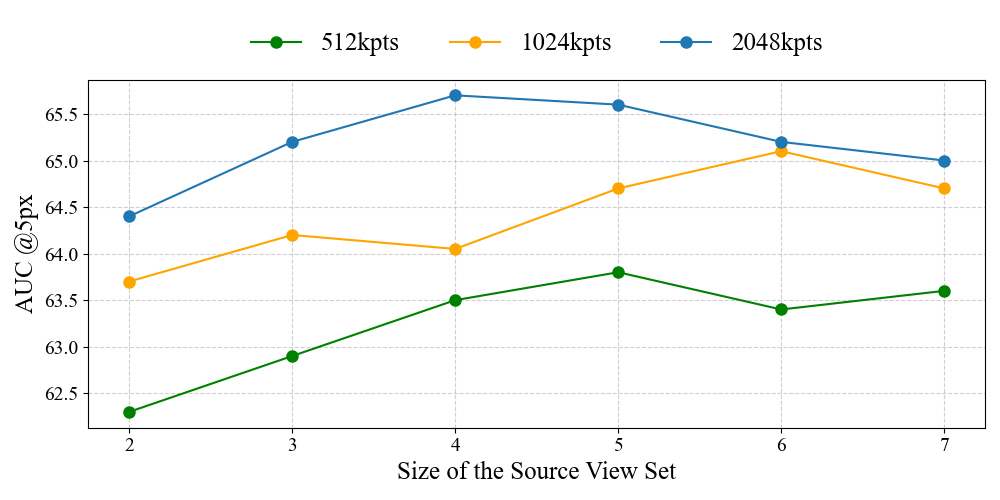}

   \caption{\textbf{Impact of the group size.} Under different numbers of keypoints, the AUC initially increases as the source view set expands, then tends to stabilize or shows a downward trend.}
   \label{fig6}
\end{figure}

\noindent\textbf{Impact of the group size.}
We evaluated the matching performance on the synthetic homography dataset~\cite{radenovic2018revisiting} by varying the size of the source view sets used as network input.
As shown in Fig.~\ref{fig6}, performance initially improves significantly as more source views are integrated into the collaborative inference process. 
However, when the number of views exceeds 5, the matching capability begins to decline. 
We hypothesize that this degradation is attributed to the expanded search space and the excessive redundant information introduced by multi-view attention, which collectively increase the learning complexity of the network.

\begin{table}
\renewcommand{\arraystretch}{1.24} 
\setlength{\tabcolsep}{2pt} 
\centering
\small
\begin{tabular}{p{0.3cm} @{\hskip -2pt} l @{\hskip -2pt} c @{\hskip -2pt} c @{\hskip -2pt} c} 
\toprule
\multirow{2}{*}{CoMatcher} & \multicolumn{2}{c}{Synthetic} & Hpatches-AUC \\ 
\cline{2-4}
& \multicolumn{1}{c}{precision} & \multicolumn{1}{c}{recall} & \multicolumn{1}{c}{@1px / @5px} \\ 
\hline
\multicolumn{1}{l}{w/o M-V feat. interaction} & \multicolumn{1}{c}{89.7} & \multicolumn{1}{c}{96.6} & 32.4 / 75.1 \\ 
\multicolumn{1}{l}{w/o atten. propagation} & \multicolumn{1}{c}{87.7} & \multicolumn{1}{c}{94.2} & 31.5 / 73.7 \\ 
\multicolumn{1}{l}{w/o M-V feat. correlation} & \multicolumn{1}{c}{90.5} & \multicolumn{1}{c}{95.3} & 33.1 / 74.9 \\ 
\hline
\multicolumn{1}{l}{full} & \multicolumn{1}{c}{\textbf{92.7}} & \multicolumn{1}{c}{\textbf{98.9}} & \textbf{34.7} / \textbf{77.1} \\ 
\bottomrule
\end{tabular}
\caption{\textbf{Ablation study on synthetic homography dataset and HPatches.} The ``full" model is the default model.}
\label{tab4}
\end{table}

\section{Conclusion}
We proposed a multi-view collaborative matching strategy to address the challenge of reliable matching uncontrolled image sets in complex scenes.
Our method, which consists of a deep collaborative matcher (CoMatcher) and a scalable groupwise pipeline, enables a holistic 3D scene understanding while inherently satisfying the cross-view projection consistency constraint.
Extensive experiments have demonstrated that exploiting inter-view connections significantly enhances matching certainty, yielding substantial benefits for downstream tasks such as pose estimation and SfM.
We hope this work encourages the research community to expand beyond pairwise matching and further explore the understanding and utilization of multi-view information.

\noindent\textbf{Acknowledgement.} 
This research is supported by NSFC projects under Grant 42471447, Development Program of China under Grant  2024YFC3811000, and the Fundamental Research Funds for the Central Universities of China under Grant 2042022dx0001.
\clearpage
\renewcommand{\thesection}{\Alph{section}} 
\renewcommand{\thesubsection}{\thesection.\arabic{subsection}} 

\setcounter{section}{0} 
\setcounter{subsection}{0} 

\maketitlesupplementary

In this supplementary material, we provide the following content:
\begin{itemize}
    \item Groupwise matching pipeline details.
    \item Images grouping algorithm.
    \item Extended methodological details.
    \item Additional experimental details.
    \item Expanded Structure-from-Motion evaluation results.
    \item Further qualitative analysis in complex scenarios.
\end{itemize}

\section{Groupwise matching pipeline details}
\begin{figure*}
    \centering
    \includegraphics[width=1\linewidth]{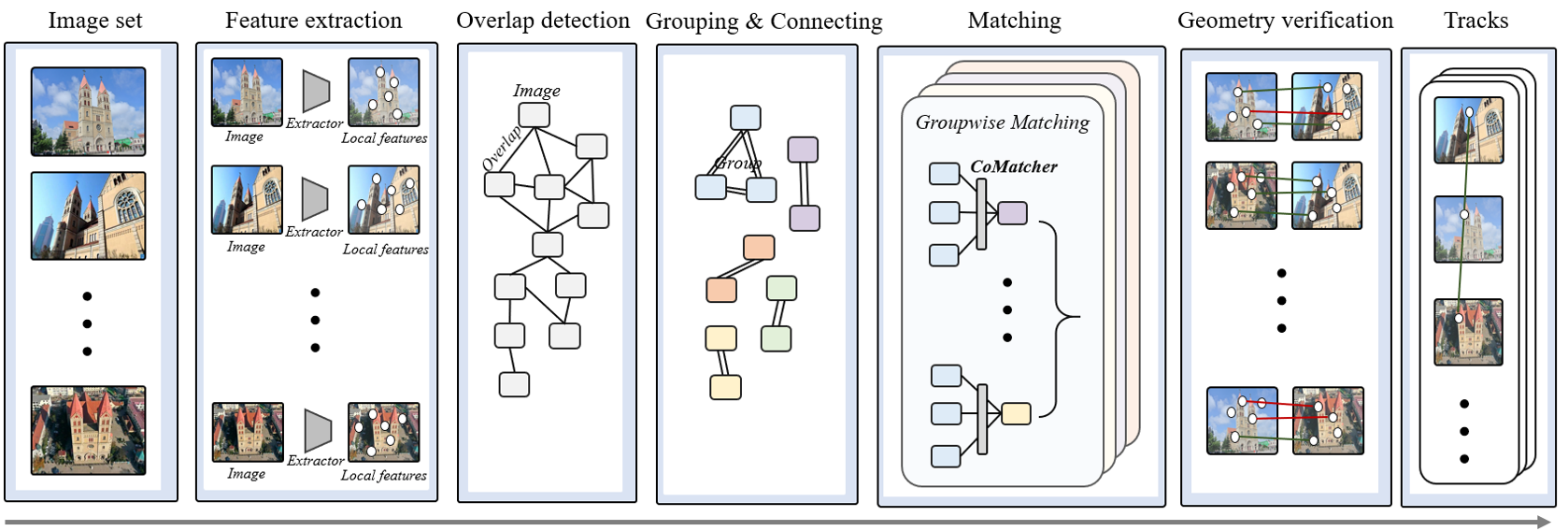}
    \caption{Overview of our groupwise matching pipeline.}
    \label{fig:Groupwise matching pipeline}
\end{figure*}

To reiterate, our method constructs a set of tracks $\{ \mathcal{M}_{k} \mid k = 1, \ldots, N_{X} \}$, where each track $\mathcal{M}_{k} = \{ \mathbf{p}^{I_{i}} \}$ corresponds to one of the $N_{X}$ 3D points observed in an unconstrained image set $\mathcal{I}= \{ I_{i} \mid i = 1, \ldots, N_{I} \}$. 
As illustrated in Fig.~\ref{fig:Groupwise matching pipeline}, the pipeline proceeds sequentially: it first extracts local features from each image, followed by identifying overlapping image pairs. 
Subsequently, the process involves grouping, connecting, and matching. 
Finally, all pairwise matches undergo robust geometric verification~\cite{fischler1981random, chum2003locally} before being merged into global tracks.
Detailed descriptions of the key components are provided in the following.

\noindent\textbf{Overlap detection.}
Overlap detection is an indispensable step in large-scale matching tasks, primarily to eliminate unnecessary matching computations for irrelevant image pairs in the dataset~\cite{agarwal2009building,jared2015reconstructing}.
Furthermore, our method requires mining complementary viewpoints from the original set.
This necessitates determining whether image pairs are co-visible and quantitatively assessing their degree of overlap.
Numerous existing retrieval methods have thoroughly investigated this problem and can perform these computations efficiently~\cite{sivic2003video, arandjelovic2016netvlad, radenovic2018fine}.
In our framework, we systematically compute pairwise co-visibility metrics to construct an overlap matrix $\mathbf{O} \in[0,1]^{N_{I} \times N_{I}}$.
Graph-theoretically, this matrix represents a weighted undirected graph $G=(V,E)$, where vertex set $V$ corresponds to images, edge set $E$ encodes co-visibility relationships, and edge weights quantify the co-visibility strength.

\begin{algorithm}[tbp]
\small
\SetAlgoInsideSkip{5pt} 
\KwIn{$V$: Node set, $O$: Edge weights, $\theta_{\min}$, $\theta_{\max}$: Thresholds, $N_{\mathcal{G}}^{\max}$: Max group size}
\KwOut{Groups $\{\mathcal{G}_1, \mathcal{G}_2, \dots\}$}

\textbf{Degree:} $d(v) = |\{u \in V \mid O_{v,u} \text{ exists}\}|$

Mark all $v \in V$ as unassigned\;

\While{exists unassigned $v \in V$}{
    $\mathcal{G} \gets \emptyset$; $i \gets \text{argmax}_{v \in V} d(v)$\;
    $\mathcal{G} \gets \mathcal{G} \cup \{i\}$; Mark $i$ as assigned; $N \gets 1$\;

    \While{$N < N_{\mathcal{G}}^{\max}$}{
        \ForEach{unassigned $v \in V$ with $O_{v,u}$ defined for some $u \in \mathcal{G}$}{
            $\text{score}(v) \gets \frac{\sum_{u \in \mathcal{G}, O_{v,u} \text{ exists}} O_{v,u}}{d(v)}$\;
        }
        $C \gets \{v \in V \mid \theta_{\min} < \text{score}(v) < \theta_{\max}, v \text{ unassigned}\}$\;
        \If{$C = \emptyset$}{\textbf{break}}
        $j \gets \text{argmax}_{v \in C} \text{score}(v)$\;
        $\mathcal{G} \gets \mathcal{G} \cup \{j\}$; Mark $j$ as assigned; $N \gets N + 1$\;
    }
}
\Return $\{\mathcal{G}_1, \mathcal{G}_2, \dots\}$\;
\caption{Images Grouping Algorithm}
\label{al:images grouping}
\end{algorithm}

\noindent\textbf{Grouping.}
In this step, we investigate the partitioning of the original image set $\mathcal{I}$ into multiple smaller groups $\{\mathcal{G}_{s} \mid s=1,  \ldots,N_{G}\}$.
This clustering problem typically requires careful design of heuristic rules for implementation.
What grouping characteristics maximize CoMatcher's collaborative matching performance?
First, pairwise co-visibility among images within each group is a necessary constraint. 
The inclusion of irrelevant images during joint estimation may cause confusion by introducing additional noise into the shared context.
Furthermore, we emphasize that group members should provide complementary rather than redundant visual information.
For example, image pairs with short baselines and similar viewpoints contribute limited additional understanding of the scene.
This requirement necessitates experimental validation to determine the optimal range of co-visibility strength.
Regarding size constraints, we enforce a maximum size $N_{G}^{\max}$ limit for each group.
This parameter is determined by the available device memory capacity, since each group must be processed as a complete input to the network. Furthermore, as demonstrated in Sec.~\ref{sec:ablation study}, incorporating excessive images does not necessarily improve performance and may even degrade results.
Based on these, we develop an iterative grouping algorithm (Sec.~\ref{sec:image grouping algorithm}).

\noindent\textbf{Connecting.}
In this step, we explicitly construct intra-group tracks to guide subsequent matching.
This can be regarded as a small unordered image set matching problem.
For implementation flexibility, we leverage existing pairwise matching frameworks through three operations:  two-view matching~\cite{sarlin2020superglue, lindenberger2023lightglue}, verification~\cite{fischler1981random} and merging.
While more robust projection positions could be obtained through a full intra-group SfM pipeline, we demonstrate that the simple approach achieves sufficient accuracy.

\noindent\textbf{Matching.}
In this step, we estimate correspondences between each image $I_{t} \subset \mathcal{I}$ and its co-visible image set $\mathcal{C}_{t}$ using a groupwise paradigm, where each $\mathcal{C}_{t}$ may represent either a single image group or the union of multiple groups.
Using CoMatcher, we process each group to establish correspondences with $I_{t}$  in a single pass, ultimately generating the complete matches for all co-visible image pairs in $\mathcal{I}$.

Our implementation follows COLMAP~\cite{schonberger2016structure} for overlap detection, geometric verification, and track merging.

\section{Images grouping algorithm}
\label{sec:image grouping algorithm}
As shown in Alg.~\ref{al:images grouping}, this algorithm iteratively searches to identify correlated image groups by utilizing both the connectivity (number of co-visible images) of each node and the edge weights.
To form a new group, we first select the un-grouped image with the highest connectivity as the seed image.
We then evaluate all co-visible images with this seed by calculating their co-visibility scores based on edge weights, filtering candidates using two predefined thresholds.
The candidate with the highest score is added to the group, and this process iterates until either no suitable candidates remain or the group reaches its maximum size.

\section{Extended methodological details}
\subsection{Architecture}
\noindent\textbf{GNN unit.}
Given the point feature $\mathbf{f}_{u}^{I_{i}}$ and a message point set $\mathcal{W}$, each GNN unit learns to integrate the message vector from $\mathcal{W}$ with $\mathbf{f}_{u}^{I_{i}}$ to update~\cite{sarlin2020superglue, lindenberger2023lightglue, sun2021loftr}:
\begin{align}
    \mathbf{f}_{u}^{I_{i}} \leftarrow \mathbf{f}_{u}^{I_{i}}+\operatorname{MLP}\left(\left[\mathbf{f}_{u}^{I_{i}} \mid \mathbf{m}_{u}^{I_{i} \leftarrow \mathcal{W}}\right]\right).
\end{align}
Here, $[\cdot \mid \cdot]$ denotes concatenation, and $\operatorname{MLP}$ represents a multi-layer perceptron.
The message vector $\mathbf{m}_{u}^{I_{i} \leftarrow \mathcal{W}}$ is computed through an attention mechanism~\cite{vaswani2017attention}, representing feature interactions between point $u$ and all points in $\mathcal{W}$.
The update $\operatorname{MLP}$ consists of a single hidden layer with a dimension of $d_h = 2d$, 
followed by a LayerNorm operation, a GeLU activation, and a linear projection 
from $(2d, d)$ with a bias term.

\noindent\textbf{Self-attention.}
CoMatcher first performs self-attention at each layer, where each point attends to all points within the same image.
For each point $u$ in $I_{i}$, the attention score is computed using a relative positional encoding scheme:
\begin{align}
a_{u v}^{I_{i} I_{i}} = \left(\mathbf{q}_{u}^{I_{i}}\right)^{\top} \mathbf{R}\left[\Delta\mathbf{p}_{u v }^{I_{i}}\right] \mathbf{k}_{v}^{I_{i}},
\end{align}
where $\mathbf{R}\left[\cdot\right]$ is a rotary encoding of the relative position between the points.

\noindent\textbf{Two-view cross-attention.}
For each source-target view pair $(I_{i}, I_{t})$, each point in $I_{i}$ attends to all points in $I_{t}$, and vice versa~\cite{sarlin2020superglue}. 
This bidirectional process is computed twice (once for each direction).
Taking a point $u$ in $I_{i}$ as an example query,  the attention scores are computed as:
\begin{align}
a_{u x}^{I_{i} I_{t}} = \left(\mathbf{q}_{u}^{I_{i}}\right)^{\top}  \mathbf{k}_{x}^{I_{t}},
\end{align}
where $\mathbf{q}_{u}^{I_{i}}$ and $\mathbf{k}_{x}^{I_{t}}$ are linearly transformed feature embeddings of the corresponding point features.
In the two-view cross-attention module, where the source view serves as the query, we incorporate the multi-view feature correlation strategy.

\noindent\textbf{Implementation of rotary encoding.}
Following~\cite{lindenberger2023lightglue}, we devide the space into $d/2$ subspaces, each rotated by an angle determined:
\begin{align}
    \mathbf{R}(\mathbf{p})=\left(\begin{array}{@{}ccc@{}}
\hat{\mathbf{R}}\left(\mathbf{b}_{1}^{\top} \mathbf{p}\right) & & 0 \\
& \ddots & \\
0 & & \hat{\mathbf{R}}\left(\mathbf{b}_{d / 2}^{\top} \mathbf{p}\right)
\end{array}\right),
\end{align}
where
\begin{align}
    \hat{\mathbf{R}}(\theta)=\left(\begin{array}{@{}cc@{}}
\cos \theta & -\sin \theta \\
\sin \theta & \cos \theta
\end{array}\right).
\end{align}
Here, $\mathbf{b}_{k}\in \mathbb{R}^{2}$ is a learned basis.

\subsection{Loss}
\noindent\textbf{Correspondence loss.}
For each pair $(I_{i}, I_{t})$, we compute ground truth matching labels $\mathcal{C}_{i, t}$ using two-view transformations (relative poses or homography), following prior works~\cite{sarlin2020superglue,sun2021loftr,lindenberger2023lightglue}.
When no other points are reprojected nearby, we label keypoints in $\mathcal{C}_{i, t}^{\emptyset}\subseteq \mathcal{W}_{i}$ or $\mathcal{C}_{t, i}^{\emptyset}\subseteq \mathcal{W}_{t}$ as non-matches, where $\mathcal{W}_{i}$ and $\mathcal{W}_{t}$ denote the feature point indices in $I_{i}$ and $I_{t}$ respectively.
We minimize the negative log-likelihood of the assignment matrix:
{\small
\begin{align}
\mathcal{L}_{\text{corr}}(I_{i},I_{t}) = 
& -\frac{1}{|\mathcal{C}_{i, t}|} \sum_{(u, x) \in \mathcal{C}_{i, t}} \log \mathbf{P}(u,x) \nonumber \\
& - \frac{1}{2|\mathcal{C}_{i, t}^{\emptyset}|} \sum_{u \in \mathcal{C}_{i, t}^{\emptyset}} \log \left(1 - \sigma_{u}^{I_{i}}\right) \nonumber \\
& - \frac{1}{2|\mathcal{C}_{t, i}^{\emptyset}|} \sum_{x \in \mathcal{C}_{t, i}^{\emptyset}} \log \left(1 - \sigma_{x}^{I_{t}}\right).
\end{align}
}
This loss function balances positive and negative samples.

\noindent\textbf{Ground-truth label of confidence loss.}
The confidence of each point in source views is quantified as the consistency probability between its correspondence estimated at the current layer and the final estimation.
The ground truth label indicates whether these two estimations are consistent, where the final estimation corresponds to the results produced by the matching head after threshold-based filtering.
To compute matching results at intermediate layers, we apply dual-softmax to the intermediate features, followed by mutual nearest-neighbor matching and threshold filtering. 
This lightweight computation effectively supervises the confidence estimator's learning while maintaining computational efficiency.

\section{Additional experimental details}
\subsection{Training details}
\noindent\textbf{Pre-training on synthetic homography dataset.}
Following~\cite{sarlin2020superglue, lindenberger2023lightglue}, we first pre-train CoMatcher on synthetic homographies of real images.
We use 150k images from the Oxford-Paris 1M distractors dataset~\cite{radenovic2018revisiting} for training.
For each training sample, we apply four distinct homography transformations to generate quadruplets consisting of three source views and one target view. 
The homographies are created by randomly sampling four image corners within each quarter of the image while ensuring a convex enclosed area to avoid degeneracies. 
We further apply random rotations and translations, keeping the corners within image boundaries to produce extreme perspective changes without border artifacts.
Each image undergoes photometric augmentation before being resized to $640\times480$ during interpolation. 
We consider correspondences with $\leq3$px symmetric reprojection error as inliers, while points failing this threshold are treated as outliers.
For feature extraction, we use 512 keypoints for SuperPoint~\cite{detone2018superpoint} and 1024 keypoints for DISK~\cite{tyszkiewicz2020disk}.
\begin{figure*}
    \centering
    \includegraphics[width=0.85\linewidth]{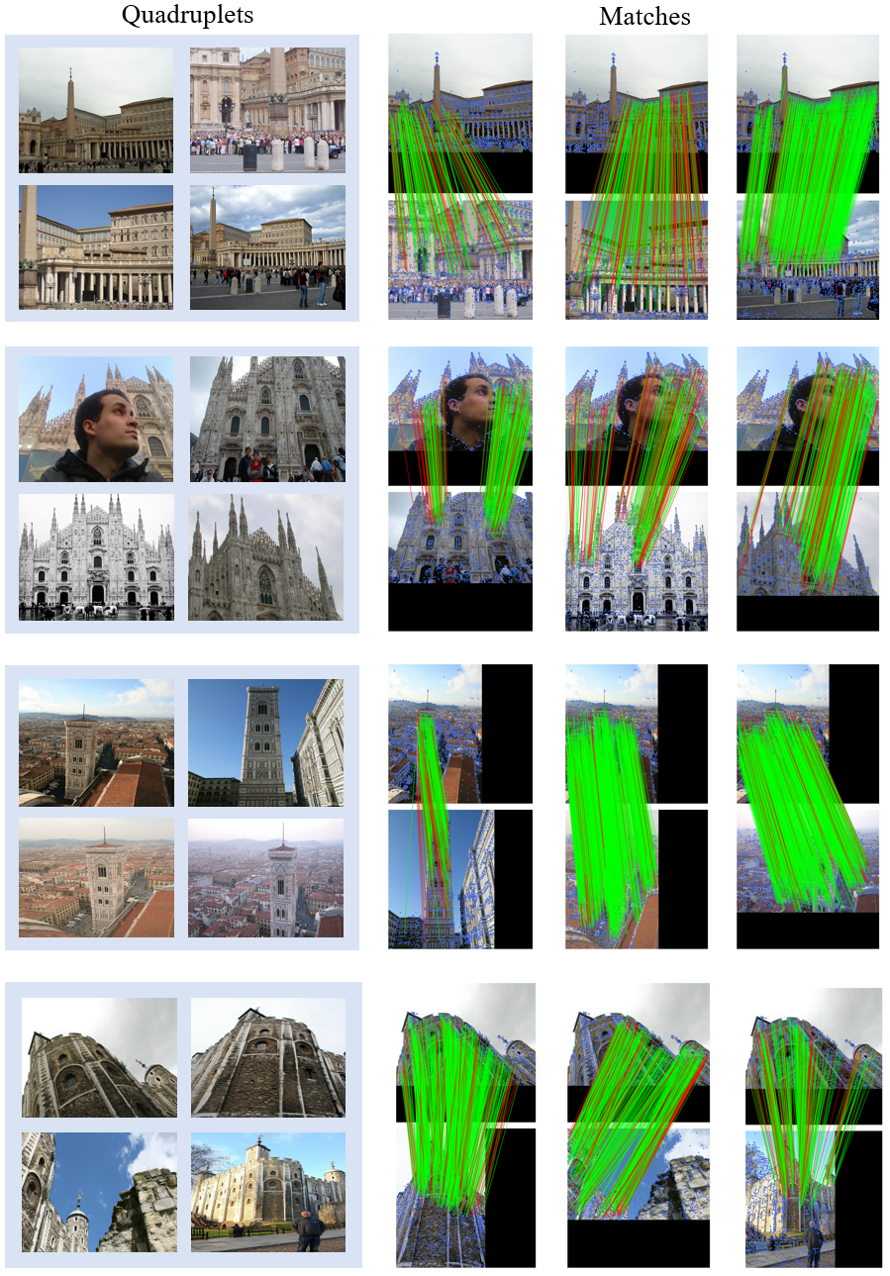}
    \caption{\textbf{Qualitative results of CoMatcher under challenging case.} In each quadruplet, the top-left image is the target view.}
    \label{fig:qualitative}
\end{figure*}

\noindent\textbf{Finetuning on MegaDepth.}
The model is fine-tuned of MegaDepth~\cite{li2018megadepth} with pseudo ground-truth camera poses and depth images.
We sample 200 co-visible multi-view quadruplets per scene, randomly select one image as the target view for training.
The sampling process follows image pair co-visibility scores, with a designed strategy to ensure balanced distribution across score intervals [0.1, 0.3], [0.3, 0.5], and [0.5, 0.9].
Images are resized such that their longer edge is 1024 pixels and zero-padded to a $1024\times1024$ resolution.
Correspondences with a reprojection error $\leq 3$ pixels and mutual nearest neighbors are labeled as inliers, 
while those with a reprojection error $> 5$ pixels are labeled as outliers. 
Points without depth or without a correspondence having a sampson error $\leq 3$ pixels are also marked as outliers. 
We extract 2048 keypoints per image.

\subsection{Evaluation details}
\noindent\textbf{Homography estimation.}
HPatches provides an ideal benchmark for evaluating our 1-to-$N$ approach.
For each scene, it supplies ground truth homographies between a target view $I_{0}$ and source views $\{I_{1}, \ldots, I_{5}\}$.
We employ a single forward pass of CoMatcher to obtain matches for all five image pairs. While End2End~\cite{roessle2023end2end} computes all pairwise correspondences in one pass for small image sets, we extract only the five required pairs for evaluation. For traditional two-view matching methods, we perform five separate forward passes to compute matches in a pairwise manner.
Following~\cite{sun2021loftr,lindenberger2023lightglue}, we resize images to a maximum edge length of 480 pixels. We evaluate homography accuracy using the mean absolute corner distance from ground truth. For each method, we fine-tune RANSAC's~\cite{fischler1981random} inlier threshold and report optimal performance.

\noindent\textbf{Relative pose estimation.}
From MegaDepth, we sampled 750 co-visible quadruplets per scene, with difficulty balanced by visual overlap following prior work~\cite{sun2021loftr,lindenberger2023lightglue,wang2024efficient}.
Each quadruplet contains one target view, yielding 4500 image pairs in total. We match each target view with its corresponding source images.
For each quadruplet, We use a single forward pass for CoMatcher and End2End~\cite{roessle2023end2end}, and perform three forward passes for two-view matching methods.
We extract 2048 local features per image, resizing each image so its larger dimension is 1600 pixels. 
For dense methods~\cite{sun2021loftr,truong2021learning, leroy2024grounding, wang2024dust3r}, we maintain the original configurations from their respective papers.

\noindent\textbf{Additional implementation details.}
In our primary experiments, we set the maximum group size to 4.
The co-visibility thresholds for group formation are denoted as $\theta_{\min}$ = 0.3 and $\theta_{\max}$= 0.7.
The choice of intra-group matching method is flexible; by default, we adopt LightGlue~\cite{lindenberger2023lightglue}.

\section{Expanded SfM evaluation results}
We further evaluate our method for SfM on both the MegaDepth~\cite{li2018megadepth} and ETH-COLMAP benchmarks~\cite{schonberger2017comparative}.
Unlike previous evaluations, this analysis focuses on two key metrics: (1) the number of landmarks (NL) and (2) the track length (TL), representing the average number of observations per landmark.
Using SuperPoint~\cite{detone2018superpoint} for local feature extraction, we compare our method with two baseline matching approaches: nearest neighbor with mutual check (NN+mutual) and LightGlue~\cite{lindenberger2023lightglue}.

For evaluation, we select one test scene from MegaDepth and two smaller scenes from the ETH-COLMAP benchmark~\cite{schonberger2017comparative}.
From each scene, we sample 50 images with sparse viewpoints and extract 2048 keypoints per image.
All matches obtained from different methods are then reconstructed using COLMAP~\cite{schonberger2016structure} with default settings.

\begin{table}[htbp]
\centering
\setlength{\tabcolsep}{6pt} 
\renewcommand{\arraystretch}{1.2} 
\begin{tabular}{clcc} 
\toprule
\textbf{Scene}              & \textbf{Method}              & \textbf{NL}    & \textbf{TL}    \\ 
\midrule
\multirow{3}{*}{Sacre Coeur} 
    & SuperPoint+NN                 & 18.1k          & 6.57           \\ 
    & SuperPoint+LightGlue          & 18.7k          & 6.92           \\ 
    & SuperPoint+\textbf{CoMatcher} & \textbf{19.3k} & \textbf{7.21}  \\ 
\midrule
\multirow{3}{*}{Fountain} 
    & SuperPoint+NN                 & 11.9k          & 4.62           \\ 
    & SuperPoint+LightGlue          & 12.6k          & 5.08           \\ 
    & SuperPoint+\textbf{CoMatcher} & \textbf{13.1k} & \textbf{5.31}  \\ 
\midrule
\multirow{3}{*}{Herzjesu} 
    & SuperPoint+NN                 & 10.4k          & 3.82           \\ 
    & SuperPoint+LightGlue          & 11.7k          & 4.12           \\ 
    & SuperPoint+\textbf{CoMatcher} & \textbf{12.4k} & \textbf{4.37}  \\ 
\bottomrule
\end{tabular}
\caption{\textbf{Structure-from-Motion on MegaDepth and ETH-COLMAP.} We report the number of landmarks (NL) and the track length (TL) of reconstruction.}
\label{tab:sfm}
\end{table}

Our method demonstrates superior performance in both landmark quantity and track length compared to conventional two-view matching approaches (See Tab.~\ref{tab:sfm}). 
This improvement directly results from CoMatcher's ability to simultaneously leverage multi-view feature information and enforce geometric consistency across views, thereby producing globally optimal correspondences that significantly enhance the reliability of track reconstruction.

\section{Further qualitative results}
To further demonstrate the robustness of our approach, we present qualitative evaluations of CoMatcher under challenging scenarios (See Fig.~\ref{fig:qualitative}). 
The proposed method demonstrates exceptional capability in precise occlusion reasoning, offering a novel solution to the long-standing challenge of wide-baseline matching in complex scenes.

{
    \small
    \bibliographystyle{ieeenat_fullname}
    \bibliography{main}

\begin{thebibliography}{60}
\providecommand{\natexlab}[1]{#1}
\providecommand{\url}[1]{\texttt{#1}}
\expandafter\ifx\csname urlstyle\endcsname\relax
  \providecommand{\doi}[1]{doi: #1}\else
  \providecommand{\doi}{doi: \begingroup \urlstyle{rm}\Url}\fi

\bibitem[Agarwal et~al.(2009)Agarwal, Snavely, Simon, Seitz, and Szeliski]{agarwal2009building}
Sameer Agarwal, Noah Snavely, Ian Simon, Steven~M Seitz, and Richard Szeliski.
\newblock Building Rome in a day.
\newblock In \emph{Proceedings of the IEEE/CVF International Conference on Computer Vision}, pages 72--79. IEEE, 2009.

\bibitem[Arandjelovic et~al.(2016)Arandjelovic, Gronat, Torii, Pajdla, and Sivic]{arandjelovic2016netvlad}
Relja Arandjelovic, Petr Gronat, Akihiko Torii, Tomas Pajdla, and Josef Sivic.
\newblock NetVLAD: CNN architecture for weakly supervised place recognition.
\newblock In \emph{Proceedings of the IEEE/CVF Conference on Computer Vision and Pattern Recognition}, pages 5297--5307, 2016.

\bibitem[Balntas et~al.(2017)Balntas, Lenc, Vedaldi, and Mikolajczyk]{balntas2017hpatches}
Vassileios Balntas, Karel Lenc, Andrea Vedaldi, and Krystian Mikolajczyk.
\newblock HPatches: A benchmark and evaluation of handcrafted and learned local descriptors.
\newblock In \emph{Proceedings of the IEEE/CVF Conference on Computer Vision and Pattern Recognition}, pages 5173--5182, 2017.

\bibitem[Barroso-Laguna et~al.(2024)Barroso-Laguna, Munukutla, Prisacariu, and Brachmann]{barroso2024matching}
Axel Barroso-Laguna, Sowmya Munukutla, Victor~Adrian Prisacariu, and Eric Brachmann.
\newblock Matching 2D images in 3D: Metric relative pose from metric correspondences.
\newblock In \emph{Proceedings of the IEEE/CVF Conference on Computer Vision and Pattern Recognition}, pages 4852--4863, 2024.

\bibitem[Bay et~al.(2006)Bay, Tuytelaars, and Van~Gool]{bay2006surf}
Herbert Bay, Tinne Tuytelaars, and Luc Van~Gool.
\newblock SURF: Speeded up robust features.
\newblock In \emph{European Conference on Computer Vision}, pages 404--417. Springer, 2006.

\bibitem[Cai et~al.(2023)Cai, Tung, Wang, Averbuch-Elor, Hariharan, and Snavely]{cai2023doppelgangers}
Ruojin Cai, Joseph Tung, Qianqian Wang, Hadar Averbuch-Elor, Bharath Hariharan, and Noah Snavely.
\newblock Doppelgangers: Learning to disambiguate images of similar structures.
\newblock In \emph{Proceedings of the IEEE/CVF International Conference on Computer Vision}, pages 34--44, 2023.

\bibitem[Chen et~al.(2021)Chen, Luo, Zhang, Zhou, Bai, Hu, Tai, and Quan]{chen2021learning}
Hongkai Chen, Zixin Luo, Jiahui Zhang, Lei Zhou, Xuyang Bai, Zeyu Hu, Chiew-Lan Tai, and Long Quan.
\newblock Learning to match features with seeded graph matching network.
\newblock In \emph{Proceedings of the IEEE/CVF International Conference on Computer Vision}, pages 6301--6310, 2021.

\bibitem[Chen et~al.(2022)Chen, Luo, Zhou, Tian, Zhen, Fang, Mckinnon, Tsin, and Quan]{chen2022aspanformer}
Hongkai Chen, Zixin Luo, Lei Zhou, Yurun Tian, Mingmin Zhen, Tian Fang, David Mckinnon, Yanghai Tsin, and Long Quan.
\newblock ASpanFormer: Detector-free image matching with adaptive span transformer.
\newblock In \emph{European Conference on Computer Vision}, pages 20--36. Springer, 2022.

\bibitem[Chum et~al.(2003)Chum, Matas, and Kittler]{chum2003locally}
Ond{\v{r}}ej Chum, Ji{\v{r}}{\'\i} Matas, and Josef Kittler.
\newblock Locally optimized RANSAC.
\newblock In \emph{Pattern Recognition}, pages 236--243. Springer, 2003.

\bibitem[DeTone et~al.(2018)DeTone, Malisiewicz, and Rabinovich]{detone2018superpoint}
Daniel DeTone, Tomasz Malisiewicz, and Andrew Rabinovich.
\newblock SuperPoint: Self-supervised interest point detection and description.
\newblock In \emph{Proceedings of the IEEE/CVF Conference on Computer Vision and Pattern Recognition Workshops}, pages 224--236, 2018.

\bibitem[Doersch et~al.(2022)Doersch, Gupta, Markeeva, Recasens, Smaira, Aytar, Carreira, Zisserman, and Yang]{doersch2022tap}
Carl Doersch, Ankush Gupta, Larisa Markeeva, Adria Recasens, Lucas Smaira, Yusuf Aytar, Joao Carreira, Andrew Zisserman, and Yi Yang.
\newblock TAP-Vid: A benchmark for tracking any point in a video.
\newblock \emph{Advances in Neural Information Processing Systems}, 35:\penalty0 13610--13626, 2022.

\bibitem[Doersch et~al.(2023)Doersch, Yang, Vecerik, Gokay, Gupta, Aytar, Carreira, and Zisserman]{doersch2023tapir}
Carl Doersch, Yi Yang, Mel Vecerik, Dilara Gokay, Ankush Gupta, Yusuf Aytar, Joao Carreira, and Andrew Zisserman.
\newblock TAPIR: Tracking any point with per-frame initialization and temporal refinement.
\newblock In \emph{Proceedings of the IEEE/CVF International Conference on Computer Vision}, pages 10061--10072, 2023.

\bibitem[Dusmanu et~al.(2019)Dusmanu, Rocco, Pajdla, Pollefeys, Sivic, Torii, and Sattler]{dusmanu2019d2}
Mihai Dusmanu, Ignacio Rocco, Tomas Pajdla, Marc Pollefeys, Josef Sivic, Akihiko Torii, and Torsten Sattler.
\newblock D2-Net: A trainable cnn for joint description and detection of local features.
\newblock In \emph{Proceedings of the IEEE/CVF Conference on Computer Vision and Pattern Recognition}, pages 8092--8101, 2019.

\bibitem[Edstedt et~al.(2024{\natexlab{a}})Edstedt, B{\"o}kman, Wadenb{\"a}ck, and Felsberg]{edstedt2024dedode}
Johan Edstedt, Georg B{\"o}kman, M{\aa}rten Wadenb{\"a}ck, and Michael Felsberg.
\newblock DeDoDe: Detect, don’t describe—Describe, don’t detect for local feature matching.
\newblock In \emph{International Conference on 3D Vision}, pages 148--157. IEEE, 2024{\natexlab{a}}.

\bibitem[Edstedt et~al.(2024{\natexlab{b}})Edstedt, Sun, B{\"o}kman, Wadenb{\"a}ck, and Felsberg]{edstedt2024roma}
Johan Edstedt, Qiyu Sun, Georg B{\"o}kman, M{\aa}rten Wadenb{\"a}ck, and Michael Felsberg.
\newblock RoMa: Robust dense feature matching.
\newblock In \emph{Proceedings of the IEEE/CVF Conference on Computer Vision and Pattern Recognition}, pages 19790--19800, 2024{\natexlab{b}}.

\bibitem[Elbayad et~al.(2020)Elbayad, Gu, Grave, and Auli]{elbayad2020depth}
Maha Elbayad, Jiatao Gu, Edouard Grave, and Michael Auli.
\newblock Depth-adaptive Transformer.
\newblock In \emph{International Conference on Learning Representations}, pages 1--14, 2020.

\bibitem[Fischler and Bolles(1981)]{fischler1981random}
Martin~A Fischler and Robert~C Bolles.
\newblock Random sample consensus: a paradigm for model fitting with applications to image analysis and automated cartography.
\newblock \emph{In Communications of the ACM}, 24\penalty0 (6):\penalty0 381--395, 1981.

\bibitem[Frahm et~al.(2010)Frahm, Fite-Georgel, Gallup, Johnson, Raguram, Wu, Jen, Dunn, Clipp, Lazebnik, et~al.]{frahm2010building}
Jan-Michael Frahm, Pierre Fite-Georgel, David Gallup, Tim Johnson, Rahul Raguram, Changchang Wu, Yi-Hung Jen, Enrique Dunn, Brian Clipp, Svetlana Lazebnik, et~al.
\newblock Building Rome on a cloudless day.
\newblock In \emph{European Conference on Computer Vision}, pages 368--381. Springer, 2010.

\bibitem[Harley et~al.(2022)Harley, Fang, and Fragkiadaki]{harley2022particle}
Adam~W Harley, Zhaoyuan Fang, and Katerina Fragkiadaki.
\newblock Particle video revisited: Tracking through occlusions using point trajectories.
\newblock In \emph{European Conference on Computer Vision}, pages 59--75. Springer, 2022.

\bibitem[Hartley and Zisserman(2003)]{hartley2003multiple}
Richard Hartley and Andrew Zisserman.
\newblock \emph{Multiple view geometry in computer vision}.
\newblock Cambridge university press, 2003.

\bibitem[He et~al.(2024)He, Sun, Wang, Peng, Huang, Bao, and Zhou]{he2024detector}
Xingyi He, Jiaming Sun, Yifan Wang, Sida Peng, Qixing Huang, Hujun Bao, and Xiaowei Zhou.
\newblock Detector-free structure-from-motion.
\newblock In \emph{Proceedings of the IEEE/CVF Conference on Computer Vision and Pattern Recognition}, pages 21594--21603, 2024.

\bibitem[Humenberger et~al.(2020)Humenberger, Cabon, Guerin, Morat, Leroy, Revaud, Rerole, Pion, de~Souza, and Csurka]{humenberger2020robust}
Martin Humenberger, Yohann Cabon, Nicolas Guerin, Julien Morat, Vincent Leroy, J{\'e}r{\^o}me Revaud, Philippe Rerole, No{\'e} Pion, Cesar de Souza, and Gabriela Csurka.
\newblock Robust image retrieval-based visual localization using kapture.
\newblock \emph{arXiv preprint arXiv:2007.13867}, 2020.

\bibitem[Jared et~al.(2015)Jared, Schonberger, Dunn, and Frahm]{jared2015reconstructing}
Heinly Jared, Johannes~L Schonberger, Enrique Dunn, and Jan-Michael Frahm.
\newblock Reconstructing the world in six days.
\newblock In \emph{Proceedings of the IEEE/CVF Conference on Computer Vision and Pattern Recognition}, 2015.

\bibitem[Jiang et~al.(2024)Jiang, Karpur, Cao, Huang, and Araujo]{jiang2024omniglue}
Hanwen Jiang, Arjun Karpur, Bingyi Cao, Qixing Huang, and Andr{\'e} Araujo.
\newblock OmniGlue: Generalizable feature matching with foundation model guidance.
\newblock In \emph{Proceedings of the IEEE/CVF Conference on Computer Vision and Pattern Recognition}, pages 19865--19875, 2024.

\bibitem[Jin et~al.(2021)Jin, Mishkin, Mishchuk, Matas, Fua, Yi, and Trulls]{jin2021image}
Yuhe Jin, Dmytro Mishkin, Anastasiia Mishchuk, Jiri Matas, Pascal Fua, Kwang~Moo Yi, and Eduard Trulls.
\newblock Image matching across wide baselines: From paper to practice.
\newblock \emph{International Journal of Computer Vision}, 129\penalty0 (2):\penalty0 517--547, 2021.

\bibitem[Karaev et~al.(2024)Karaev, Rocco, Graham, Neverova, Vedaldi, and Rupprecht]{karaev2024cotracker}
Nikita Karaev, Ignacio Rocco, Benjamin Graham, Natalia Neverova, Andrea Vedaldi, and Christian Rupprecht.
\newblock CoTracker: It is better to track together.
\newblock In \emph{European Conference on Computer Vision}, pages 18--35. Springer, 2024.

\bibitem[Leroy et~al.(2024)Leroy, Cabon, and Revaud]{leroy2024grounding}
Vincent Leroy, Yohann Cabon, and J{\'e}r{\^o}me Revaud.
\newblock Grounding image matching in 3D with MASt3R.
\newblock In \emph{European Conference on Computer Vision}, pages 71--91. Springer, 2024.

\bibitem[Li and Snavely(2018)]{li2018megadepth}
Zhengqi Li and Noah Snavely.
\newblock MegaDepth: Learning single-view depth prediction from internet photos.
\newblock In \emph{Proceedings of the IEEE/CVF Conference on Computer Vision and Pattern Recognition}, pages 2041--2050, 2018.

\bibitem[Lindenberger et~al.(2021)Lindenberger, Sarlin, Larsson, and Pollefeys]{lindenberger2021pixel}
Philipp Lindenberger, Paul-Edouard Sarlin, Viktor Larsson, and Marc Pollefeys.
\newblock Pixel-perfect structure-from-motion with featuremetric refinement.
\newblock In \emph{Proceedings of the IEEE/CVF International Conference on Computer Vision}, pages 5987--5997, 2021.

\bibitem[Lindenberger et~al.(2023)Lindenberger, Sarlin, and Pollefeys]{lindenberger2023lightglue}
Philipp Lindenberger, Paul-Edouard Sarlin, and Marc Pollefeys.
\newblock LightGlue: Local feature matching at light speed.
\newblock In \emph{Proceedings of the IEEE/CVF International Conference on Computer Vision}, pages 17627--17638, 2023.

\bibitem[Lowe(2004)]{lowe2004distinctive}
David~G Lowe.
\newblock Distinctive image features from scale-invariant keypoints.
\newblock \emph{International Journal of Computer Vision}, 60:\penalty0 91--110, 2004.

\bibitem[Ma et~al.(2021)Ma, Jiang, Fan, Jiang, and Yan]{ma2021image}
Jiayi Ma, Xingyu Jiang, Aoxiang Fan, Junjun Jiang, and Junchi Yan.
\newblock Image matching from handcrafted to deep features: A survey.
\newblock \emph{International Journal of Computer Vision}, 129\penalty0 (1):\penalty0 23--79, 2021.

\bibitem[Maset et~al.(2017)Maset, Arrigoni, and Fusiello]{maset2017practical}
Eleonora Maset, Federica Arrigoni, and Andrea Fusiello.
\newblock Practical and efficient multi-view matching.
\newblock In \emph{Proceedings of the IEEE/CVF International Conference on Computer Vision}, pages 4568--4576, 2017.

\bibitem[Mur-Artal et~al.(2015)Mur-Artal, Montiel, and Tardos]{mur2015orb}
Raul Mur-Artal, Jose Maria~Martinez Montiel, and Juan~D Tardos.
\newblock ORB-SLAM: A versatile and accurate monocular SLAM system.
\newblock \emph{In IEEE Transactions on Robotics}, 31\penalty0 (5):\penalty0 1147--1163, 2015.

\bibitem[Olsson and Enqvist(2011)]{olsson2011stable}
Carl Olsson and Olof Enqvist.
\newblock Stable structure-from-motion for unordered image collections.
\newblock In \emph{Image Analysis}, pages 524--535. Springer, 2011.

\bibitem[Pan et~al.(2024)Pan, Bar{\'a}th, Pollefeys, and Sch{\"o}nberger]{pan2024global}
Linfei Pan, D{\'a}niel Bar{\'a}th, Marc Pollefeys, and Johannes~L Sch{\"o}nberger.
\newblock Global structure-from-motion revisited.
\newblock In \emph{European Conference on Computer Vision}, 2024.

\bibitem[Radenovi{\'c} et~al.(2018{\natexlab{a}})Radenovi{\'c}, Iscen, Tolias, Avrithis, and Chum]{radenovic2018revisiting}
Filip Radenovi{\'c}, Ahmet Iscen, Giorgos Tolias, Yannis Avrithis, and Ond{\v{r}}ej Chum.
\newblock Revisiting Oxford and Paris: Large-scale image retrieval benchmarking.
\newblock In \emph{Proceedings of the IEEE/CVF Conference on Computer Vision and Pattern Recognition}, pages 5706--5715, 2018{\natexlab{a}}.

\bibitem[Radenovi{\'c} et~al.(2018{\natexlab{b}})Radenovi{\'c}, Tolias, and Chum]{radenovic2018fine}
Filip Radenovi{\'c}, Giorgos Tolias, and Ond{\v{r}}ej Chum.
\newblock Fine-tuning CNN image retrieval with no human annotation.
\newblock \emph{IEEE Transactions on Pattern Analysis and Machine Intelligence}, 41\penalty0 (7):\penalty0 1655--1668, 2018{\natexlab{b}}.

\bibitem[Revaud et~al.(2019)Revaud, De~Souza, Humenberger, and Weinzaepfel]{revaud2019r2d2}
Jerome Revaud, Cesar De~Souza, Martin Humenberger, and Philippe Weinzaepfel.
\newblock R2D2: Reliable and repeatable detector and descriptor.
\newblock \emph{Advances in Neural Information Processing Systems}, 32, 2019.

\bibitem[Roessle and Nie{\ss}ner(2023)]{roessle2023end2end}
Barbara Roessle and Matthias Nie{\ss}ner.
\newblock End2end multi-view feature matching with differentiable pose optimization.
\newblock In \emph{Proceedings of the IEEE/CVF International Conference on Computer Vision}, pages 477--487, 2023.

\bibitem[Rublee et~al.(2011)Rublee, Rabaud, Konolige, and Bradski]{rublee2011orb}
Ethan Rublee, Vincent Rabaud, Kurt Konolige, and Gary Bradski.
\newblock ORB: An efficient alternative to SIFT or SURF.
\newblock In \emph{Proceeding of the IEEE/CVF International Conference on Computer Vision}, pages 2564--2571. Ieee, 2011.

\bibitem[Sarlin et~al.(2019)Sarlin, Cadena, Siegwart, and Dymczyk]{sarlin2019coarse}
Paul-Edouard Sarlin, Cesar Cadena, Roland Siegwart, and Marcin Dymczyk.
\newblock From coarse to fine: Robust hierarchical localization at large scale.
\newblock In \emph{Proceedings of the IEEE/CVF Conference on Computer Vision and Pattern Recognition}, pages 12716--12725, 2019.

\bibitem[Sarlin et~al.(2020)Sarlin, DeTone, Malisiewicz, and Rabinovich]{sarlin2020superglue}
Paul-Edouard Sarlin, Daniel DeTone, Tomasz Malisiewicz, and Andrew Rabinovich.
\newblock SuperGlue: Learning feature matching with graph neural networks.
\newblock In \emph{Proceedings of the IEEE/CVF Conference on Computer Vision and Pattern Recognition}, pages 4938--4947, 2020.

\bibitem[Sattler et~al.(2017)Sattler, Torii, Sivic, Pollefeys, Taira, Okutomi, and Pajdla]{sattler2017large}
Torsten Sattler, Akihiko Torii, Josef Sivic, Marc Pollefeys, Hajime Taira, Masatoshi Okutomi, and Tomas Pajdla.
\newblock Are large-scale 3d models really necessary for accurate visual localization?
\newblock In \emph{Proceedings of the IEEE/CVF Conference on Computer Vision and Pattern Recognition}, pages 1637--1646, 2017.

\bibitem[Schonberger and Frahm(2016)]{schonberger2016structure}
Johannes~L Schonberger and Jan-Michael Frahm.
\newblock Structure-from-Motion revisited.
\newblock In \emph{Proceedings of the IEEE/CVF Conference on Computer Vision and Pattern Recognition}, pages 4104--4113, 2016.

\bibitem[Schonberger et~al.(2017)Schonberger, Hardmeier, Sattler, and Pollefeys]{schonberger2017comparative}
Johannes~L Schonberger, Hans Hardmeier, Torsten Sattler, and Marc Pollefeys.
\newblock Comparative evaluation of hand-crafted and learned local features.
\newblock In \emph{Proceedings of the IEEE/CVF Conference on Computer Vision and Pattern Recognition}, pages 1482--1491, 2017.

\bibitem[Schuster et~al.(2022)Schuster, Fisch, Gupta, Dehghani, Bahri, Tran, Tay, and Metzler]{schuster2022confident}
Tal Schuster, Adam Fisch, Jai Gupta, Mostafa Dehghani, Dara Bahri, Vinh Tran, Yi Tay, and Donald Metzler.
\newblock Confident adaptive language modeling.
\newblock \emph{Advances in Neural Information Processing Systems}, 35:\penalty0 17456--17472, 2022.

\bibitem[Sivic and Zisserman(2003)]{sivic2003video}
Sivic and Zisserman.
\newblock Video Google: A text retrieval approach to object matching in videos.
\newblock In \emph{Proceedings in the IEEE/CVF International Conference on Computer Vision}, pages 1470--1477. IEEE, 2003.

\bibitem[Su et~al.(2024)Su, Ahmed, Lu, Pan, Bo, and Liu]{su2024roformer}
Jianlin Su, Murtadha Ahmed, Yu Lu, Shengfeng Pan, Wen Bo, and Yunfeng Liu.
\newblock RoFormer: Enhanced transformer with rotary position embedding.
\newblock \emph{Neurocomputing}, 568:\penalty0 127063, 2024.

\bibitem[Sun et~al.(2021)Sun, Shen, Wang, Bao, and Zhou]{sun2021loftr}
Jiaming Sun, Zehong Shen, Yuang Wang, Hujun Bao, and Xiaowei Zhou.
\newblock LoFTR: Detector-free local feature matching with transformers.
\newblock In \emph{Proceedings of the IEEE/CVF Conference on Computer Vision and Pattern Recognition}, pages 8922--8931, 2021.

\bibitem[Sun et~al.(2023)Sun, Yu, Tao, Li, Tang, and Qian]{sun2023unified}
Kun Sun, Jinhong Yu, Wenbing Tao, Xin Li, Chang Tang, and Yuhua Qian.
\newblock A unified feature-spatial cycle consistency fusion framework for robust image matching.
\newblock \emph{Information Fusion}, 97:\penalty0 101810, 2023.

\bibitem[Truong et~al.(2021)Truong, Danelljan, Van~Gool, and Timofte]{truong2021learning}
Prune Truong, Martin Danelljan, Luc Van~Gool, and Radu Timofte.
\newblock Learning accurate dense correspondences and when to trust them.
\newblock In \emph{Proceedings of the IEEE/CVF Conference on Computer Vision and Pattern Recognition}, pages 5714--5724, 2021.

\bibitem[Tyszkiewicz et~al.(2020)Tyszkiewicz, Fua, and Trulls]{tyszkiewicz2020disk}
Micha{\l} Tyszkiewicz, Pascal Fua, and Eduard Trulls.
\newblock DISK: Learning local features with policy gradient.
\newblock \emph{Advances in Neural Information Processing Systems}, 33:\penalty0 14254--14265, 2020.

\bibitem[Vaswani et~al.(2017)Vaswani, Shazeer, Parmar, Uszkoreit, Jones, Gomez, Kaiser, and Polosukhin]{vaswani2017attention}
Ashish Vaswani, Noam Shazeer, Niki Parmar, Jakob Uszkoreit, Llion Jones, Aidan~N Gomez, {\L}ukasz Kaiser, and Illia Polosukhin.
\newblock Attention is all you need.
\newblock \emph{Advances in Neural Information Processing Systems}, 30, 2017.

\bibitem[Wang et~al.(2024{\natexlab{a}})Wang, Leroy, Cabon, Chidlovskii, and Revaud]{wang2024dust3r}
Shuzhe Wang, Vincent Leroy, Yohann Cabon, Boris Chidlovskii, and Jerome Revaud.
\newblock DUSt3R: Geometric 3D vision made easy.
\newblock In \emph{Proceedings of the IEEE/CVF Conference on Computer Vision and Pattern Recognition}, pages 20697--20709, 2024{\natexlab{a}}.

\bibitem[Wang et~al.(2024{\natexlab{b}})Wang, He, Peng, Tan, and Zhou]{wang2024efficient}
Yifan Wang, Xingyi He, Sida Peng, Dongli Tan, and Xiaowei Zhou.
\newblock Efficient LoFTR: Semi-dense local feature matching with sparse-like speed.
\newblock In \emph{Proceedings of the IEEE/CVF Conference on Computer Vision and Pattern Recognition}, pages 21666--21675, 2024{\natexlab{b}}.

\bibitem[Xiao et~al.(2024)Xiao, Wang, Zhang, Xue, Peng, Shen, and Zhou]{xiao2024spatialtracker}
Yuxi Xiao, Qianqian Wang, Shangzhan Zhang, Nan Xue, Sida Peng, Yujun Shen, and Xiaowei Zhou.
\newblock SpatialTracker: Tracking any 2D pixels in 3D space.
\newblock In \emph{Proceedings of the IEEE/CVF Conference on Computer Vision and Pattern Recognition}, pages 20406--20417, 2024.

\bibitem[Xu et~al.(2024)Xu, Chen, Xu, Wang, Lu, and Guo]{xu2024local}
Shibiao Xu, Shunpeng Chen, Rongtao Xu, Changwei Wang, Peng Lu, and Li Guo.
\newblock Local feature matching using deep learning: A survey.
\newblock \emph{Information Fusion}, 107:\penalty0 102344, 2024.

\bibitem[Yi et~al.(2016)Yi, Trulls, Lepetit, and Fua]{yi2016lift}
Kwang~Moo Yi, Eduard Trulls, Vincent Lepetit, and Pascal Fua.
\newblock LIFT: Learned invariant feature transform.
\newblock In \emph{European Conference on Computer Vision}, pages 467--483. Springer, 2016.

\bibitem[Zhang et~al.(2010)Zhang, Dong, Jia, Wong, and Bao]{zhang2010efficient}
Guofeng Zhang, Zilong Dong, Jiaya Jia, Tien-Tsin Wong, and Hujun Bao.
\newblock Efficient non-consecutive feature tracking for structure-from-Motion.
\newblock In \emph{European Conference on Computer Vision}, pages 422--435. Springer, 2010.

\end{thebibliography}
}

\end{document}